\documentclass[lettersize,journal]{IEEEtran}
\usepackage{amsmath,amsfonts}
\usepackage{algorithmic}
\usepackage{algorithm}
\usepackage{array}
\usepackage[caption=false,font=normalsize,labelfont=sf,textfont=sf]{subfig}
\usepackage{textcomp}
\usepackage{stfloats}
\usepackage{url}
\usepackage{verbatim}
\usepackage{graphicx}
\usepackage{cite}
\usepackage{tcolorbox}
\usepackage{hyperref}
\usepackage{amsmath}

\usepackage{times}
\usepackage{soul}
\usepackage{url}
\usepackage[utf8]{inputenc}
\usepackage[small]{caption}
\usepackage{graphicx}
\usepackage{amsmath}
\usepackage{amsthm}
\usepackage{tcolorbox}

\usepackage{booktabs}
\usepackage{algorithm}
\usepackage{algorithmic}
\usepackage[switch]{lineno}
\usepackage{amsmath}
\usepackage{multirow}
\usepackage{xcolor}

\usepackage{pifont}
\newcommand{\cmark}{\ding{51}} 
\newcommand{\xmark}{\ding{55}} 

\usepackage{makecell}
\usepackage{graphicx}

\usepackage{hyperref}

\begin{document}

\title{STHMoE: Hypergraph-Enhanced Heterogeneous Dependency Coordination for LLM-Based Urban Traffic Data Forecasting}


\author{
Jiawen Chen~$^*$, Qi Shao~$^*$, Yongjian Chang, Mingtong Zhou, Duxin Chen$^\dagger$, Wenwu Yu$^\dagger$,~\IEEEmembership{Senior Member, IEEE,} 
\thanks{This research was supported by the National Key Research and Development Program of China (Grant No.G2025YFF0524100), the National Natural Science Foundation of China (Grants No.62233004, 62273090, and T2541017), the Jiangsu Provincial Scientific Research Center of Applied Mathematics (Grant No.BK20233002), and the Basic Research Program of Jiangsu (Grants No.BK20253018 and BK20253020).}%
\thanks{Jiawen Chen, Qi Shao, Mingtong Zhou, Duxin Chen, Wenwu Yu are with School of Mathematics, Southeast University, Nanjing 210096, China. Jiangsu Province Center for Applied Mathematical Sciences, Nanjing 210096, China. (chenjiawen@seu.edu.cn; shaoqi@seu.edu.cn; 213221863@seu.edu.cn; chendx@seu.edu.cn; wwyu@seu.edu.cn). Jiawen Chen and Qi Shao equally contribute to this work. Corresponding authors: Duxin Chen, Wenwu Yu.}
\thanks{Yongjian Chang is with School of Cyber Science and Engineering, Southeast University, Nanjing 210096, China. (yongjianchang@seu.edu.cn).}}
\markboth{Manuscript (under review)}%
{Shell \MakeLowercase{\textit{et al.}}: A Sample Article Using IEEEtran.cls for IEEE Journals}


\maketitle

\begin{abstract}
Spatio-temporal traffic forecasting is a fundamental big data analytics task for intelligent transportation systems, where massive urban sensor streams exhibit heterogeneous, non-stationary, and structurally dynamic patterns. Although recent deep learning and large language model (LLM)-based methods have advanced traffic forecasting, they often remain temporally centered and lack effective coordination of temporal, spectral, pairwise spatial, and higher-order structural cues under evolving traffic regimes.
To address this heterogeneous dependency coordination problem, we propose STHMoE, a Spatio-Temporal Hypergraph-Enhanced Mixture of Experts framework for urban traffic data forecasting. STHMoE decouples traffic dynamics into frequency-domain, time-domain, spatio-domain, and higher-order spatial representations, which are modeled by prompt-guided heterogeneous experts built upon a partially frozen LLM backbone. The first three experts leverage domain-specific statistical prompts, while the higher-order spatio expert uses a structural placeholder prompt and obtains dependency information from an adaptive hypergraph module. To capture evolving spatial structures in traffic data,, STHMoE jointly learns first-order graph dependencies and higher-order group interactions without predefined topologies. An entropy-aware MoE router with coefficient-of-variation load balancing adaptively fuses expert outputs while improving expert utilization and routing confidence. Experiments on 10 real-world traffic benchmarks show that STHMoE achieves competitive performance against temporal, spatio-temporal graph, and LLM-based baselines. Code and datasets are available at {\url{https://github.com/jiawenchen10/STHMoE}}.
\end{abstract}


\section{Introduction}

\begin{figure}[htbp]
  \centering
  \setlength{\belowcaptionskip}{-15pt}
  \includegraphics[width=0.95\linewidth]{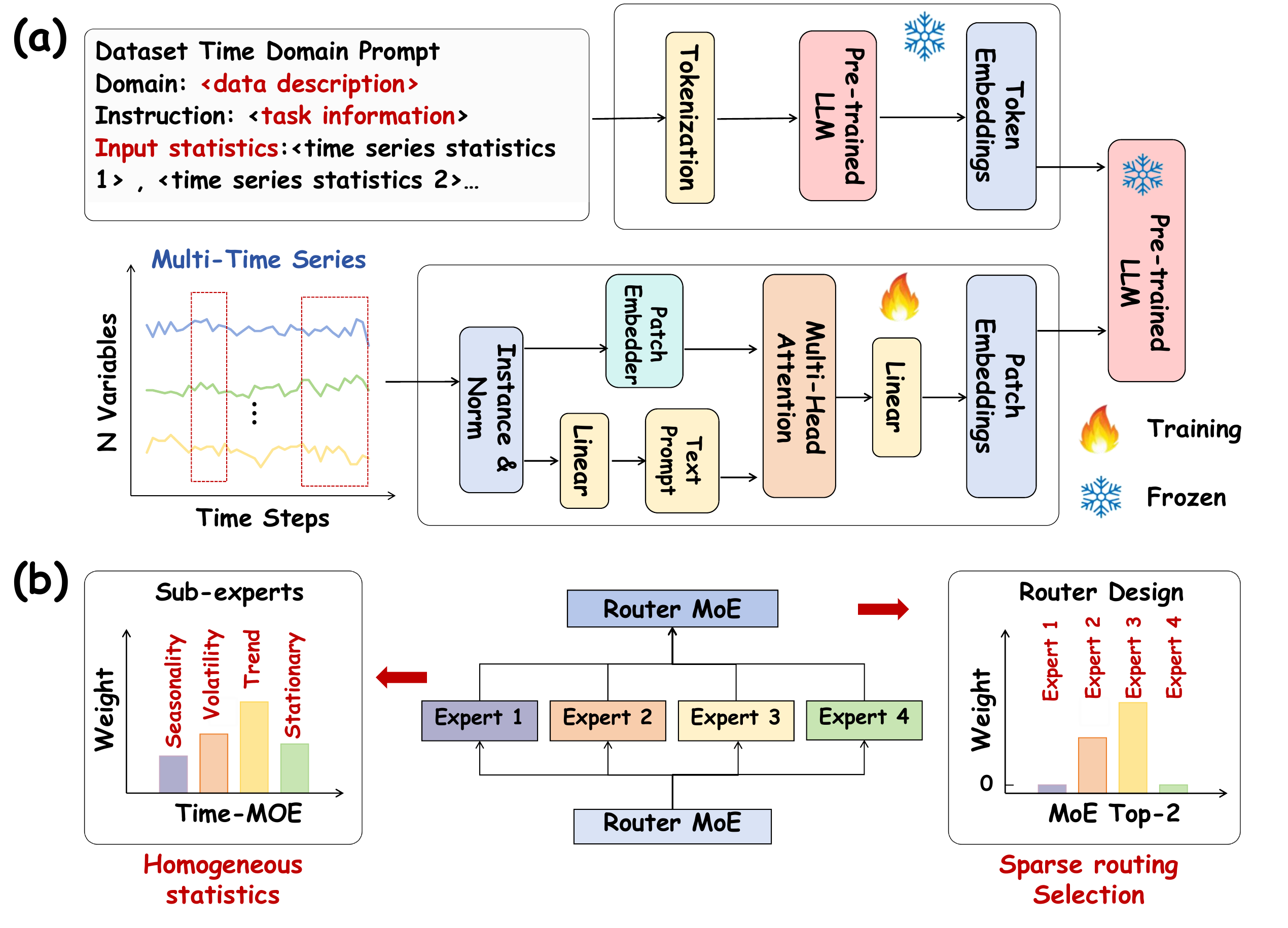}
  \caption{LLM-method design. (a) LLM-based methods (e.g., Time-LLM~\cite{jin2024time}) inject rich time context, instruction, timestamp, statistics. (b) Time-series MoEs (e.g., Time-MoE~\cite{xiaoming2025time}) introduce Trend, Seasonality, Volatility, Stationary subexperts. It employs Top-K routing. }
  \label{fig02}
\end{figure}

\IEEEPARstart{S}{patio-temporal} prediction serves as a fundamental component of modern data-driven decision-making, with applications in urban traffic forecasting~\cite{wang2020deep,10896813}, climate modeling~\cite{bi2023accurate}, energy grid management~\cite{jin2024survey}, and weather prediction~\cite{10473200}. The deployment of Intelligent Transportation Systems makes spatio-temporal analysis essential for congestion mitigation, route optimization~\cite{yu2024systems}, and proactive resource allocation. However, such large-scale traffic data are inherently  heterogeneous~\cite{chen2025decoupling}, multi-source~\cite{OpenCity}, exhibiting evolving temporal patterns, spatial correlations, and traffic regimes across regions and time periods. These characteristics call for scalable and adaptive forecasting frameworks that can exploit massive observational data for complex urban systems.

In recent years, large language models (LLMs) have offered new insights to capture underlying statistics and semantic information in traffic data  forecasting~\cite{jin2024time,xiaoming2025time,liu2024time,li2024gpt23,zhou2023one} in Fig.~\ref{fig02}~(a). Despite their promising progress, existing LLM-based forecasting methods are still predominantly designed around temporal representations, with limited capacity to characterize frequency periodicities, dynamic spatial correlations, and structured topological dependencies. Recent studies have attempted to enhance LLM-based forecasting by incorporating graph neural networks (GNNs) for spatial dependency modeling~\cite{zhou2023one,shang2026multiscale,liu2025st}. However, many of these hybrid frameworks still build spatial relations from predefined or static geographic structures~\cite{yuan2024unist}, which may be insufficient for traffic systems whose heterogeneous dependencies evolve across time, regions, and traffic regimes. Consequently, how to jointly coordinate temporal statistics, frequency patterns, adaptive spatial correlations, and structural topology information within an LLM-based forecasting framework~\cite{jin2024time,chen2025decoupling,liu2026semanticenhanced} remains under-explored, particularly when such heterogeneous cues can be integrated to adapt to evolving traffic regimes.

\begin{figure}[htbp]
\centering
\setlength{\belowcaptionskip}{-15pt}
\includegraphics[width=0.95\linewidth]{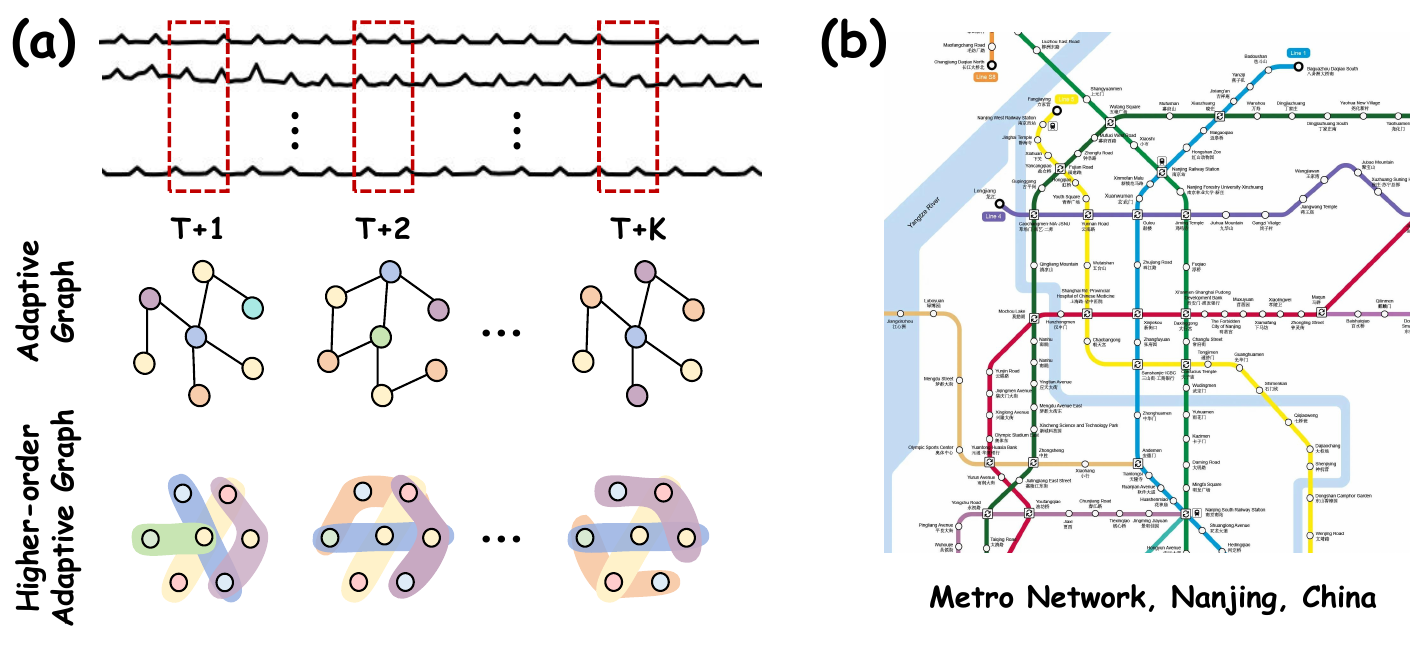}
\caption{Motivation for Higher-order spatio-temporal modeling in traffic flow forecasting. (a)
Adaptive graphs capture only station-to-station edges, adaptive hypergraphs express group-wise co-activations among multiple stations. (b) Higher-order dependencies:  Nanjing metro network, with cross-line transfers and functional zones.}
\label{fig01}
\end{figure}

To capture the heterogeneity of traffic patterns, Mixture-of-Experts (MoE) architectures have recently been introduced into time-series and spatio-temporal forecasting. As illustrated in Fig.~\ref{fig02}(b), existing MoE-based methods usually assign different experts to specific temporal or spectral patterns. For example, TimeMoE~\cite{xiaoming2025time} routes time-series predictors according to temporal characteristics, FreqMoE~\cite{liu2025freqmoe} decouples signals into frequency-aware components, and ST-MoE~\cite{li2023st} focuses on localized spatio-temporal features. Despite their effectiveness, these models usually emphasize specific representational views, and their ability to coordinate temporal dynamics, frequency periodicities, pairwise spatial relations, and higher-order spatial interactions remains limited.
Moreover, the commonly used Top-K routing strategy~\cite{deepseekai2024deepseekv32,li2025uni} discards unselected expert outputs and may over-concentrate routing weights on a few dominant experts. Under rare congestion events or distribution shifts, such hard sparsification can suppress complementary expert responses, thereby weakening performance and robustness to extreme traffic variations.

 Early spatio-temporal forecasting methods commonly rely on GNNs with predefined adjacency matrices~\cite{gan2024novel, zheng2023spatio}, which limits their ability to adapt to evolving topological patterns~\cite{yu2018spatio}. In real-world systems, inter-node influences continuously change due to evolving dynamics~\cite{chen2025decoupling} and external triggers, such as weather events~\cite{10473200} or demand surges~\cite{zheng2023soup}. Recent adaptive graph learning methods alleviate this limitation by capturing time-varying pairwise relationships~\cite{11077392,wu2020connecting}. However, they still represent spatial interactions through binary edges. As illustrated in Fig.~\ref{fig01}, a single congestion event may affect multiple downstream regions and induce group-wise co-activations among related stations. Such multi-way dependencies are difficult to characterize using pairwise graphs alone, motivating the need for higher-order spatial modeling.


To address these challenges, we propose Spatio-Temporal Hypergraph-enhanced Mixture-of-Experts (STHMoE), an LLM-based heterogeneous dependency coordination framework for traffic data forecasting. STHMoE is designed to coordinate temporal, spectral, pairwise spatial, and higher-order structural cues while adaptively learning dynamic first-order and higher-order dependencies. The main contributions are summarized as follows:

\begin{itemize}
    \item We propose STHMoE, a heterogeneous expert learning framework that integrates LLM-based representation modeling with MoE-based adaptive fusion for spatio-temporal forecasting. The framework decouples traffic dynamics into frequency-domain, time-domain, pairwise spatial, and higher-order spatial experts, enabling coordinated modeling of complementary dependency patterns.
    
    \item We design a prompt-guided expert adaptation strategy to bridge traffic semantics and pre-trained language models. The frequency, time, and pairwise spatial experts are guided by domain-specific statistical prompts, while the higher-order spatial expert adopts a structural placeholder prompt, with its higher-order dependency information explicitly encoded by the adaptive hypergraph module.
    
    \item We develop an adaptive structure-enhanced module that jointly learns adaptive higher-order hypergraph interactions without relying on fixed geographic topologies. This module captures both dynamic pairwise correlations and group-wise co-activation patterns, improving the representation of evolving spatial dependencies.
    
    \item We conduct extensive experiments across diverse real-world traffic benchmarks. The empirical results show that STHMoE achieves competitive performance over spatio-temporal graph models and recent LLM-based baselines.
\end{itemize}

\section{RELATED WORK}
In this section, we review the existing literature on spatio-temporal forecasting and the adaptation of large language models for traffic prediction.

\subsection{Spatio-Temporal Prediction} 
Early deep learning approaches to spatio-temporal forecasting jointly model spatial and temporal components. Recent advances increasingly combine advanced temporal sequence modeling with graph-structured spatial learning. Modern time-series architectures, such as DLinear~\cite{zeng2023transformers}, TimesNet~\cite{wu2023timesnet}, PatchTST~\cite{nietime}, and iTransformer~\cite{liu2023itransformer}, achieve strong performance by capturing multi-scale temporal patterns. However, they typically neglect the complex spatial structures inherent in real-world traffic systems. To incorporate spatial dependencies, several methods integrate graph neural networks with sequence models. For instance, STG-NCDE~\cite{choi2022graph} combines neural controlled differential equations with graph representations to capture continuous-time dynamics, while STJGCN~\cite{zheng2023spatio} jointly attends to spatial and temporal dimensions. Furthermore, hypergraph-based architectures, including STHSepNet~\cite{chen2025decoupling} and GPT-ST~\cite{li2024gpt23}, enhance representational expressiveness by capturing higher-order spatial interactions and topological drift. Frequency-aware variants, such as FEDformer~\cite{zhou2022fedformer} and Autoformer~\cite{wu2021autoformer}, improve the handling of temporal non-stationarity but lack adaptive spatial modeling capabilities. Despite these advances, most existing approaches rely on static or predefined geographic structures, restricting their adaptability to evolving relationships and non-stationary distributions. 
In contrast, STHMoE builds on temporal, frequency-aware, adaptive graph, and hypergraph modeling, but differs by coordinating these heterogeneous cues as specialized experts to jointly capture dynamic pairwise and higher-order spatial dependencies.

\subsection{Large Language Models for Spatio-Temporal Prediction}
Large language models feature massive parameter spaces and strong generalization capabilities~\cite{liang2025foundation}, leading to their increasing adoption in time-series analysis~\cite{ma2024survey} and forecasting tasks~\cite{liu2024unitime,chang2025llm4ts}. By leveraging large-scale pre-training, LLM-based approaches demonstrate promising capabilities in modeling long-range temporal dependencies. Because text-based LLMs cannot directly process numerical time-series data, prior research focuses on bridging this representation gap. Early frameworks reformulate forecasting as next-token prediction by encoding numerical values into text, achieving competitive zero-shot performance~\cite{gruver2023large}. Subsequent methods improve computational efficiency via specialized encodings and cross-domain alignment, utilizing bimodal representations~\cite{liu2025timecma} and lightweight multimodal designs~\cite{liu2024unitime,lin2025sparsetsf,chen2025decoupling}. More general techniques, such as prompt reprogramming~\cite{jin2024time}, contrastive embedding~\cite{li2024unicl}, and parameter-efficient fine-tuning~\cite{chang2023llm4ts,zhou2023one}, further align the numerical and textual spaces. Beyond pure temporal modeling, recent extensions attempt to incorporate spatial context. ST-LLM~\cite{liu2025st} treats each spatial location as a distinct token using partially frozen attention layers, while UrbanGPT~\cite{li2024urbangpt} adapts LLMs to urban forecasting via external knowledge injection. Nevertheless, most existing LLM-based methods primarily emphasize temporal statistics and lack explicit mechanisms for dynamic spatial dependency learning. STHMoE addresses this bottleneck through a heterogeneous expert coordination mechanism that aligns temporal, spectral, pairwise spatial, and higher-order structural cues at fine-grained spatio-temporal coordinates, rather than simply combining an LLM with graph or hypergraph modules.

\section{Preliminaries}

In this section, we introduce the preliminaries of spatio-temporal traffic forecasting. For clarity, the primary notations utilized throughout this paper are summarized in Table~\ref{tab:notations}.

\subsection{Problem Formulation}

Let the spatial structure be represented by a weighted graph $G = (V, E, A)$, where $V$ is the set of $N$ nodes, $E$ denotes the set of edges, and $A \in \mathbb{R}^{N \times N}$ is the predefined adjacency matrix. Given the historical observations $X_{t-L+1:t} \in \mathbb{R}^{L \times N \times F}$, where $L$ is the look-back window and $F$ represents the feature dimension, spatio-temporal forecasting aims to predict the future values over the subsequent $H$ steps, denoted as $\hat{X}_{t+1:t+H} \in \mathbb{R}^{H \times N \times F}$. 
The objective is to learn a mapping function $f_\theta$ that estimates the future $H$ time steps:
\begin{equation} 
[x_{t-L+1}, x_{t-L+2}, \dots, x_t, G, \Phi] \xrightarrow{f_\theta(\cdot)} [\hat{x}_{t+1}, \dots, \hat{x}_{t+H}],
\end{equation}
where $\theta$ denotes the model parameters and $\Phi$ represents auxiliary prompt information used to guide the prediction. In this work, STHMoE does not rely on $A$ and instead learns adaptive first-order and higher-order structures.

\subsection{Adaptive Network Construction}

\begin{table}[htbp]
  \centering
  \caption{Summary of Key Notations}
  \renewcommand{\arraystretch}{1.2}  
  \begin{tabular}{p{0.6in}p{2.65in}}
    \toprule 
    Notation & Description \\
    \midrule 
    $X, \overline{X}$ & Input node features and spatially pooled global trend \\
    $P, S$ & Patch length and sliding stride for temporal patching \\
    $X_P, \hat{X}_P$ & Sequence of temporal patches and their embeddings \\
    $W_i^\cdot, A_i^\cdot, B_i^\cdot$ & Frozen pre-trained weights and LoRA low-rank matrices \\
    $X^{(\ell)}$ & Node features at the $\ell$-th layer of MixProp \\
    $k$ & Hyperedge order number \\
    $A_{adp}^{(k)}$ & Learnable $k$-order adaptive adjacency matrix \\
    $h_t^v, m_t^v$ & State of node $v$ at time $t$ and its aggregated features \\
    $P_i, \widetilde{O}_i$ & Output of the $i$-th expert and its fused representation \\
    $X_e^{(k)}, X_v^{(k)}$ & $k$-order hyperedge features and node features \\
    $\rho, \gamma$ & Scaling factor and first/higher-order balance parameter \\
    $R, w$ & Raw routing score tensor and normalized gating weights \\
    $H^R, Z^R$ & Number of routing heads and learnable embeddings \\
    $\tau$ & Learnable temperature parameter for expert gating \\
    $\hat{X}$ & Final prediction of the STHMoE framework \\
    \bottomrule
  \end{tabular}
  \label{tab:notations}
\end{table}

In real-world scenarios, fixed geographic networks fail to describe the dynamic coupled relations between traffic stations.
We introduce an adaptive adjacency matrix, $ \tilde{A}_{\text{adp}} $, which aims to capture adaptive dependencies between nodes. Given node features $ E_1, E_2 \in \mathbb{R}^{N \times d} $, we employ a shared-parameter feed forward neural network (FFN) to generate node embeddings, which are then mapped to $ F_1, F_2 \in \mathbb{R}^{N \times N} $ as follows: To introduce non-linearity, we construct an asymmetric adjacency matrix $ A^{(1)}_{\text{adp}} \in \mathbb{R}^{N \times N} $: 
\begin{align}
& F_1 = \tanh (\rho \cdot \mathrm{FFN} (E_1)), 
 F_2 = \tanh (\rho \cdot \mathrm{FFN} (E_2)), \\
& A^{(1)}_{\text{adp}} = \mathrm{ReLU} (\tanh (\rho \cdot (F_1 F^\top_2 - F_2 F^\top_1))).
\end{align} 
where $\rho$ is a scaling factor that modulates the saturation rate of the activation function. The discrepancy between $ F_1 $ and $ F_2 $ captures directional relationships between nodes, which leverages learned structure in the traffic system. 

\begin{figure*}
\centering
\setlength{\belowcaptionskip}{-15pt}
\includegraphics[width=0.92\linewidth]{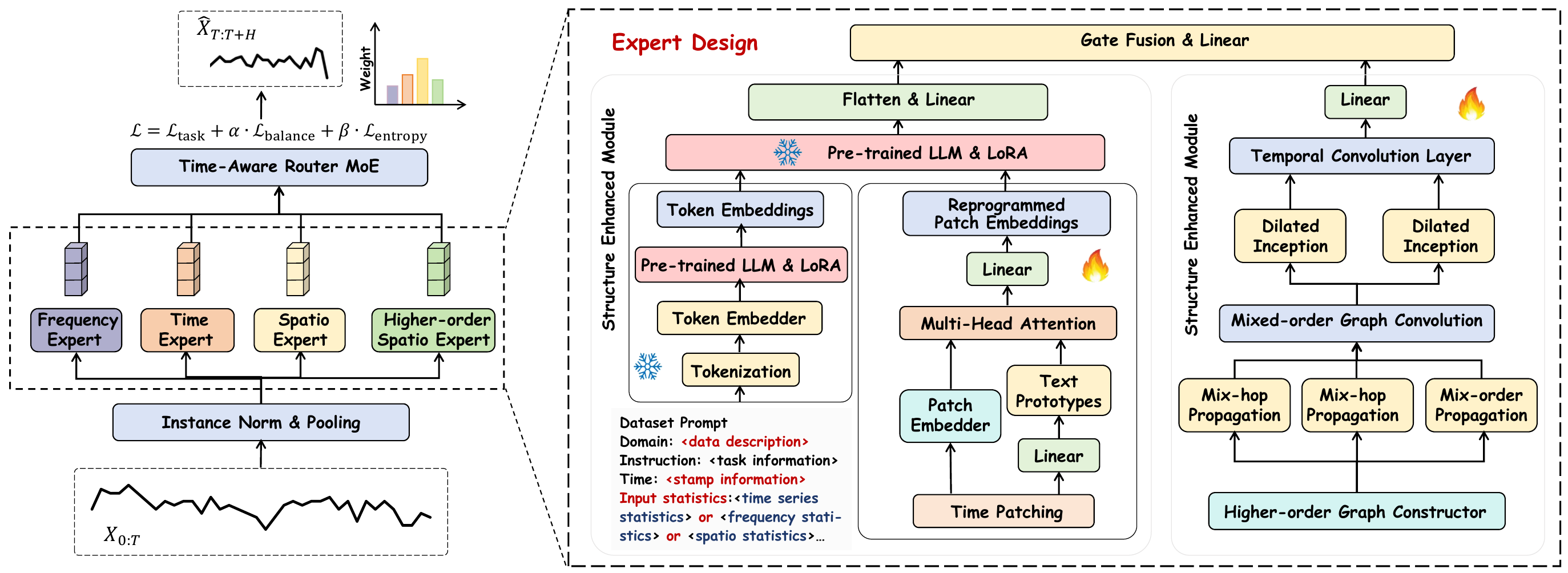}
\caption{Overall architecture of STHMoE. The framework integrates four domain-specific experts, an adaptive structure-enhanced module for pairwise and higher-order spatial modeling, and a time-aware MoE router for fine-grained expert fusion. Load-balancing and entropy regularization are used to improve expert utilization and routing confidence.}

\label{fig03}
\end{figure*}

\subsection{Adaptive Higher-order Graph Construction}
To capture higher-order spatial dependencies beyond pairwise interactions, we construct an adaptive hypergraph in the learned feature space. 
Given node features $E_3 \in \mathbb{R}^{N \times d} $, we first obtain latent feature representations with the FFN layer:
\begin{equation}
F_3 = \tanh\!\big(\rho\cdot \mathrm{FFN}(E_3)\big),
\end{equation}
Leveraging the feature embeddings $F_3 = [f_1, \dots, f_n]$, higher-order relations are adaptively constructed with $k$-nearest neighbours. For each node $v_i$, we select its \(q\) nearest neighbours in the latent feature space, denoted by $\mathcal{N}_q(v_i)$, and construct a hyperedge \(e_i=\{v_i\}\cup\mathcal{N}_q(v_i)\). The hyperedge cardinality is thus \(k=q+1\).
This adaptive hypergraph is encoded by an adjacency matrix $A^{(k)}_{\text{adp}} \in \mathbb{R}^{n \times m}$, where $n$ is the number of nodes and $m$ is the number of hyperedges. The elements are defined as: $A^{(k)}_{\text{adp}, ij}$ = 1,if $v_i \in e_j$, and 0, otherwise. 
The adaptive hypergraph mechanism captures richer and more complex higher-order group interactions within the spatio-temporal dependencies of urban traffic data.

\section{Methodology}
In this section, we provide a detailed elaboration of the proposed STHMoE and its components.

\subsection{Overview}
The proposed STHMoE framework is illustrated in Fig.~\ref{fig03}, which shows the three tightly coupled stages. First, a patch embedding module compresses historical observations into compact temporal tokens. Second, the time-, frequency-, and pairwise spatio-domain experts are guided by statistical prompts, while the higher-order spatio expert employs a structural placeholder prompt and receives high-order dependency information from the adaptive hypergraph module. Meanwhile, a structure-enhanced spatial module constructs adaptive first-order and $k$-order hypergraph adjacency matrices, injecting dynamic topological representations into each expert to enhance spatio-temporal reasoning. Third, an entropy-aware MoE routing layer aggregates these enriched outputs via temperature-scaled soft routing on the full spatio-temporal context.  Ultimately, a composite objective function jointly optimizes the framework by minimizing forecasting error, balancing expert utilization via a coefficient-of-variation loss, and sharpening routing confidence through entropy regularization.



\subsection{Patch Embedding Module}
Given the input node features
$X \in \mathbb{R}^{B \times N \times T \times F}$,
we first perform spatial pooling by applying average pooling along the node dimension
to extract global fluctuation trends shared across regions:
$
\bar{X} = \mathrm{P}(X)$, 
where $\bar{X} \in \mathbb{R}^{B \times T \times F}$.
To capture long-range temporal dependencies and improve computational efficiency,
$\bar{X}$ is segmented along the temporal dimension into a sequence of
patches by applying a sliding window of length $P$
with stride $S$. Specifically, each patch corresponds to a contiguous subsequence
of $P$ time steps, yielding $
X_P \in \mathbb{R}^{B \times N_P \times P \times F}$, 
$N_P = \left\lfloor \frac{T - P}{S} \right\rfloor + 1$.
Here, each temporal patch is treated as a token and projected into a latent space, resulting in the embedded representation $\hat{X}_P \in \mathbb{R}^{B \times N_P \times d_m}$, where $d_m$ denotes the hidden dimension of the large language model.

\subsection{Temporal Sequence Modeling via LoRA-Augmented LLM}
\paragraph{Prompt Adaptation Module}
Large language models are primarily pre-trained on large-scale textual corpora and therefore lack explicit inductive biases for traffic forecasting. To bridge this gap, we employ a cross-domain alignment strategy that transforms time-series observations into textual tokens, enabling LLMs to exploit their reasoning capabilities for forecasting tasks.
To integrate temporal, frequency, and structural knowledge from traffic data, we design four domain-specific experts to model temporal, frequency, and structural patterns. As shown in Fig.~\ref{fig:three_expert_prompts}, the time-, frequency-, and spatio-domain experts are guided by statistical prompts, whereas the higher-order spatio expert employs a structural placeholder prompt and an adaptive hypergraph module. Each prompt consists of dataset descriptions, task instructions, and statistical characteristics, and is prepended to reprogrammed patch embeddings before being fed into the LLM backbone.

\paragraph{Partially Frozen LLM Backbone}
We adopt a pre-trained large language model with most parameters frozen to model long-range temporal dependencies. The selected pre-trained model serves as the backbone of STHMoE and is composed of $N$ stacked Transformer layers. Let $z = \{z^1, z^2, \dots, z^N\}$ denote the layer-wise hidden representations, where the initial input $z^1$ is formed by concatenating prompt embeddings with time-series embeddings. At each layer $i$, the hidden state $z^i$ is first processed by multi-head self-attention (MHSA) and combined with residual connections and Layer Normalization~(LN) to produce an intermediate representation $\tilde{z}^i$. This representation is then passed through a Feed-Forward Network~(FFN) followed by Layer Normalization, yielding the output $z^{i+1}$ for the next layer. The latter can be written as:
\begin{align}
\label{eqllm}
& (Q_i, K_i, V_i) = (W_i^Q z^i, W_i^K z^i, W_i^V z^i), \\
&\mathrm{MHSA}(z^i) = W([\text{H}_1 \| \cdots \| \text{H}_h]),  
 \text{H}_j = \sigma \!\left(\frac{Q_i K_i^\top}{\sqrt{d}}\right) V_i, \label{eq08} \\
& \tilde{z}^i = \mathrm{LN}\!\big(z^i + \mathrm{MHSA}(z^i)\big), \\ & z^{i+1} = \mathrm{LN}\!\big(\tilde{z}^i + \mathrm{FFN}(\tilde{z}^i)\big), 
\end{align}
where $z^i$ denotes the hidden state at layer $i$ and $\tilde{z}^i$ represents the post-attention intermediate feature. As the LLM outputs representations at the token level, we introduce a linear projection layer to map these token embeddings to patch-level forecasting outputs. The framework is compatible with both encoder-style and decoder-style pre-trained language models. For BERT, bidirectional self-attention is adopted, whereas decoder-style backbones rely on causal self-attention.

\paragraph{LoRA-Augmented Multi-Head Self-Attention}
To adapt the frozen LLM to spatio-temporal forecasting efficiently, we introduce Low-Rank Adaptation (LoRA)~\cite{hulora} into the attention mechanism. Specifically, each query, key, and value projection is augmented by a low-rank residual update. For the $i$-th layer, the augmented projections are defined as:
\begin{equation}
\begin{aligned}
Q_i &= \big(W_i^Q + B_i^Q A_i^Q\big) z^i, \\
K_i &= \big(W_i^K + B_i^K A_i^K\big) z^i, \\
V_i &= \big(W_i^V + B_i^V A_i^V\big) z^i,
\end{aligned}
\end{equation}
where $W_i^\cdot \in \mathbb{R}^{d \times d}$ are the frozen pre-trained weight matrices, and $B_i^\cdot \in \mathbb{R}^{d \times r}$ and $A_i^\cdot \in \mathbb{R}^{r \times d}$ are the trainable low-rank matrices with rank $r \ll d$. The attention heads are then computed using the standard softmax operation and concatenated across all $h$ heads. The resulting representation is further projected by a LoRA-augmented output transformation:
\begin{equation}
\mathrm{MHSA}(z^i)
=
\big(W_i^O + B_i^O A_i^O\big)
\big[\text{Head}_1 \| \cdots \| \text{Head}_h\big],
\end{equation}
where $W_i^O \in \mathbb{R}^{d \times d}$ denotes the frozen output projection matrix, and $B_i^O, A_i^O$ are its low-rank adaptation parameters.

\subsection{Decoupled Spatio-Temporal Domain Experts}
Spatio-temporal forecasting relies on temporal evolution and spatial propagation. However, processing their heterogeneous signals within a unified network can exacerbate representation entanglement. Specifically, temporal dynamics exhibit both explicit time-domain fluctuations and implicit frequency-domain periodicities~\cite{jin2024time,liu2025freqmoe}. Similarly, spatial propagation involves localized pairwise relations~\cite{yu2018spatio,wu2020connecting} and complex higher-order group interactions~\cite{chen2025decoupling}. Because these four fundamental patterns exhibit distinct representational properties, joint modeling often leads to mutual interference and suboptimal feature extraction. Therefore, we decouple the representational space by introducing four parallel domain-specific experts (time, frequency, pairwise spatial, and higher-order spatial). This decoupled MoE design enables each expert to specialize in a distinct representational view and provides complementary information for adaptive routing.

\begin{figure}
\centering
\setlength{\belowcaptionskip}{-15pt}    
\includegraphics[width=0.95\linewidth]{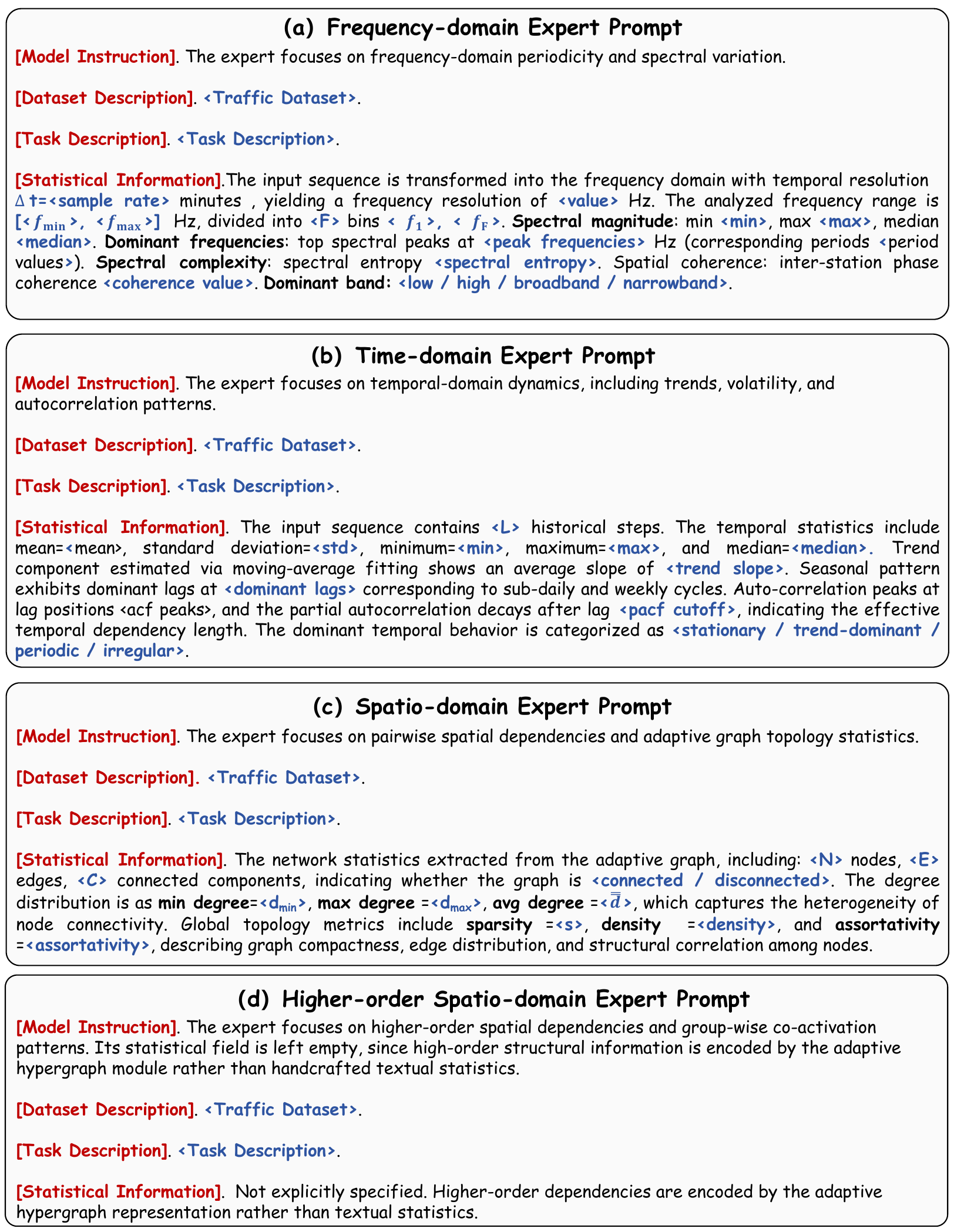}
\caption{Prompt designs for four domain-specific experts: (a) frequency-domain expert, (b) time-domain expert, (c) spatio-domain expert, and (d) higher-order spatio expert. }
\label{fig:three_expert_prompts}
\end{figure}

\paragraph{Frequency Expert}
Traffic flow data often exhibit pronounced periodicity and recurrent fluctuations, which are difficult to fully capture using purely time-domain descriptions. To capture such spectral patterns, we introduce a frequency-domain expert that operates in the transformed frequency space. Specifically, this expert leverages frequency-aware statistical cues provided by the structured Frequency Prompt in Fig.~\ref{fig:three_expert_prompts}~(a) to guide the LLM backbone in modeling dominant periodic components and multi-scale oscillatory dynamics. Based on these frequency-guided representations, the expert directly produces the time-series forecasting output $P_1$. 

\paragraph{Time Expert}
Traffic flows also involve complex, non-periodic temporal dependencies, including short-term transient fluctuations and long-term trends. To capture these time-aligned features, we deploy a time-domain expert that models temporal evolution from historical observations. To inject explicit temporal knowledge into the LLM, we construct a time prompt that encapsulates key local statistical dynamics, sequence volatility, and global trends without relying on frequency conversions, as detailed in Fig.~\ref{fig:three_expert_prompts}(b). The resulting expert prediction is denoted as $P_2$.

\paragraph{Spatio Expert}
Traffic dynamics are also rooted in complex spatial dependencies~\cite{millan2025topology} that isolated temporal models cannot resolve. We introduce a spatio expert that distills adaptive graph topologies into key network statistics to guide the LLM backbone. As depicted in Fig.~\ref{fig:three_expert_prompts}(c), the corresponding prompt describes the spatial structure from multiple perspectives, including node connectivity, graph density, sparsity, connected components, and degree heterogeneity. By exposing the LLM to these compact topological descriptors, the expert can account for both local interactions and global spatial organization and output prediction $P_3$.

\paragraph{Higher-order Spatio Expert}
To capture spatio-temporal dependencies beyond pairwise relationships, we design a higher-order spatio-temporal graph neural network as the higher-order spatio expert. As shown in Fig.~\ref{fig:three_expert_prompts}(d), this expert is assigned a structured prompt whose statistical field is left empty and used only as a structural placeholder. This module explicitly models complex interactions via an adaptive hypergraph. 
Given an adaptive hypergraph adjacency matrix $A^{(k)}_{\text{adp}}$, the module executes a two-phase information passing scheme: node-to-hyperedge and hyperedge-to-node.
In the node-to-hyperedge phase, each hyperedge $e$ accumulates information from its associated nodes. Subsequently, features of all hyperedges containing node $v$ are aggregated back to update the node representation:
 \begin{equation}
\small
\begin{aligned}
 & X^{(k)}_{e} = \sigma\big(\sum_{v\in N(e)} D_{v}^{-\frac{1}{2}}A^{(k)}_{adp,v}D_e^{-\frac{1}{2}} X_{v}^{(k-1)} W \big),  \\
& X_{v}^{(k)} = \sum_{e \in \mathcal{E}(v)} D_{e}^{-\frac{1}{2}} A^{(k),T}_{adp, e} D_{v}^{-\frac{1}{2}} X_{e}^{(k)}, \label{eq21}
\end{aligned}
\end{equation}
where $W$ is a trainable parameter matrix, $D_v\in R^{N\times N}$,$D_e \in R^{|\mathcal{E}| \times |\mathcal{E}|}$, $\sigma(\cdot)$ is the ReLU activation function, and $\mathcal{E}(v)$ indicates the set of hyperedges connected to node $v$. Let the final output of this higher-order module be $X^{(k)}_{\text{adp}} = X_{v}^{(k)}$.
To jointly exploit pairwise and higher-order spatial dependencies, we fuse the first-order features $X^{(1)}_{\text{adp}}$ and higher-order features $X^{(k)}_{\text{adp}}$ via a learnable weighted combination:
\begin{equation}
\label{eq:higherorder}
  X^{(k)}_{\text{mixed}} = \gamma \cdot X^{(1)}_{\text{adp}} + (1-\gamma) \cdot X^{(k)}_{\text{adp}},
\end{equation}
where $\gamma \in [0,1]$ is a learnable parameter that balances the contributions of higher-order spatial representations.

The fused structural representation $X^{(k)}_{\text{mixed}}$ is then fed into a stacked Spatio-Temporal Convolution Module to formulate the final expert output. This module stacks multiple ST-Blocks, each comprising a Spatio-Block and a Temporal-Block. 
In the Spatio-Block, each node state $h_{t}^{(v)}$ is initialized with the fused features, i.e., $h_{0}^{(v)} = X^{(k)}_{\text{mixed}} \in \mathbb{R}^{B \times N \times T \times F}$ and updated by aggregating features from its neighbors as 
\begin{equation}
\label{eq13}
\begin{aligned}
 h_t^{v} = \sigma ( (1 + \epsilon) h_{t-1}^{v} + m_t^{v}), \quad m_t^{v} = \sum_{u \in N (v)} h_{t-1}^{u} , \\
\end{aligned}
\end{equation}
where $ m_t^{v} $ is the aggregated neighborhood feature at time $ t $, and $ \epsilon $ is a learnable parameter. 
The Temporal-Block comprises 1-D dilated convolution layers with a gating mechanism featuring only an output gate. 
Given the input $\chi \in \mathbb{R}^{T \times N \times F}$, the gated spatio-temporal representation $h$ is defined as:
\begin{equation}
\label{eq14}
\begin{aligned}
{P}_4 = \tanh (q (\chi)) \otimes \sigma (q (\chi)) , 
\end{aligned}
\end{equation}
where $q (\chi)$ is the output of the dilated convolution layers, $\otimes$ denotes the Hadamard product, and $\sigma$ is the sigmoid function.

\paragraph{Structure-Enhanced Module.}
To learn spatial information after spatio-temporal decoupling, we introduce a mixed multi-layer aggregation mechanism (MixProp $M_{\theta}$). Given the input features $X^{(0)} \in \mathbb{R}^{B \times N \times T}$, we employ the MixProp module to iteratively update node representations. To regulate information flow, the features at the $(\ell+1)$-th layer are updated via a gating-based fusion:
\begin{equation}
\begin{aligned}
    &G^{(\ell)} = \sigma \big(W_g X^{(\ell)}\big), \\
    &X^{(\ell+1)} = G^{(\ell)} \odot X^{(\ell)} + \big(1 - G^{(\ell)}\big) \odot \big(\hat{A} X^{(\ell)}\big),
\end{aligned}
\end{equation}
where $X^{(\ell)}$ represents the node features at the $\ell$-th layer, $\hat{A} = D_v^{-\frac{1}{2}} (A^{(1)}_{\text{adp}} + I) D_v^{-\frac{1}{2}}$ is the normalized adaptive adjacency matrix with self-loops, $D$ is the degree matrix, $W_g \in \mathbb{R}^{C \times C}$ is a learnable weight matrix, $\sigma(\cdot)$ is the sigmoid activation function, and $\odot$ denotes element-wise multiplication. The MixProp module employs $K$-layer propagation to expand the receptive field and capture dependencies between distant nodes.  
To further capture bi-directional spatial dependencies, we apply this MixProp-based graph convolution $M_{\theta}$ on the adaptive adjacency matrix $A_{\text{adp}}$. These operations extract first-order and transposed first-order features, defined as:
\begin{equation}
    X_{\text{adp}}^{(1)} = M_{\theta} \big(\bar{X}, A_{\text{adp}}, K\big) + M_{\theta}\big(\bar{X}, A^{T}_{\text{adp}}, K\big),  
\end{equation}
where $\bar{X}$ represents the input time-series features, and $K$ is the number of propagation layers. This extracted adaptive graph structure information is fed into the spatio-temporal convolution module described in Eqs.~\eqref{eq13}-\eqref{eq14}, yielding the structural position embedding ($\text{PE}$).

To reintegrate the decoupled representations, we inject the high-order spatial context captured by the structure-enhanced module into the predictions from the pre-trained LLM experts. Specifically, let $P_i \in \mathbb{R}^{B \times T \times N}$ denote the preliminary output of the $i$-th domain expert ($i \in \{1,2,3,4\}$), and let $\text{PE} \in \mathbb{R}^{B \times T \times N}$ be the spatial structural embedding (with feature dimensions aligned to the LLM expert outputs). For each expert $i$, we perform an adaptive gated fusion between its temporal prediction $P_i$ and the shared spatial signal $\text{PE}$:
\begin{equation}
\begin{aligned}
  & \text{Gate}_i = \sigma\!\big( \mathrm{FFN}([P_i, \text{PE}]) \big), \\
  & \widetilde{O}_i = \mathrm{FFN}\big(P_i \odot \text{Gate}_i + \text{PE} \odot (1 - \text{Gate}_i)\big),
\end{aligned}
\end{equation}
where $[\cdot, \cdot]$ denotes channel-wise concatenation, $\odot$ denotes element-wise multiplication. The resulting fused representations $\widetilde{O}_i$ preserve both the specific reasoning of each independent expert and the global heterogeneous spatial structure. These enriched representations are subsequently forwarded to the entropy-aware MoE routing layer for final aggregation.

\subsection{Spatio-Temporal Hypergraph Mixture-of-Experts}
To adaptively integrate heterogeneous predictive cues, we introduce a spatio-temporal hypergraph-enhanced mixture-of-experts module. Unlike conventional MoE architectures that route experts at the sequence or token level, STHMoE performs coordinate-wise routing over temporal steps and spatial nodes, enabling the model to handle temporal non-stationarity and spatial heterogeneity in a fine-grained manner.

\paragraph{Time-Aware Router.} 
Let $\tilde{O}_{e}\in\mathbb{R}^{B\times T\times N}$ denote the prediction of the $e$-th expert, where $B$, $T$, and $N$ are the batch size, temporal length, and number of spatial nodes, respectively. Given the spatio-temporal context, the router generates a raw score tensor $R\in\mathbb{R}^{B\times T\times N\times E}$:
\begin{equation}
    R_{b,t,n,e} = \sum_{j=1}^{H^R} W_{b,t,n,j}^R \odot \operatorname{ReLU} \left( Q_{b,t,n,j}^R (Z_{e}^R)^\top \right),
\label{eq:routing_matrix}
\end{equation}
where $H^R$ is the number of routing heads, $Q_{j}^R$ and $W_{j}^R$ are the query and weight projections for the $j$-th head, and $Z^R \in \mathbb{R}^{E \times d_r}$ contains the learnable embeddings for all experts, here $d_r$ is routing embedding dimension.

\paragraph{Expert Gate} 
To construct a differentiable expert mixture, the scores $R$ are normalized along the expert dimension using a temperature-scaled soft routing into a normalized gating tensor $W \in \mathbb{R}^{B \times T \times N \times E}$, it can be denoted as
\begin{equation}
    W_{b,t,n,e} = \frac{\exp(R_{b,t,n,e} / \tau)}{\sum_{l=1}^{E} \exp(R_{b,t,n,l} / \tau)},
\label{eq:gating_matrix}
\end{equation}
where $\tau$ a learnable temperature parameter that controls the sharpness of expert assignment, here $\tau$ is initialized to 1.0.

\paragraph{Adaptive Expert Fusion}
The final prediction $\hat{Y}$ is obtained by aggregating the outputs of all heterogeneous experts, which specialize in the time domain, frequency domain, pairwise spatial relations, and higher-order spatial interactions. The element-wise fusion $\hat{X}\in \mathbb{R}^{H \times N \times F}$ is formulated as:
\begin{equation}
    \hat{X} = \sum_{i=1}^{4} \left( W_{:,:,:,i} \odot \tilde{O}_i \right).
\label{eq:expert_fusion}
\end{equation}
This operation combines representations from different heterogeneous domains at each spatio-temporal coordinate.

\subsection{Training Objective}
The STHMoE model is optimized by jointly minimizing the forecasting error and regularizing the expert routing distribution. The total training loss is formulated as a weighted sum of the task-specific loss, a load-balancing auxiliary loss and entropy regularization loss:
\begin{equation}
    \mathcal{L}_{\text{total}} = \mathcal{L}_{\text{task}} + \alpha \cdot \mathcal{L}_{\text{balance}} + \beta \cdot \mathcal{L}_{\text{entropy}},
\label{eq:total_loss}
\end{equation}
where $\alpha$ and $\beta$ control the contributions of the load-balancing and entropy regularization terms, respectively. We employ Mean Absolute Error (MAE) as the primary task loss:  
\begin{equation}
\mathcal{L}_{task} = \frac{1}{B H N} \sum_{b=1}^{B} \sum_{t=1}^{H} \sum_{n=1}^{N} \left| X_{b,t,n} - \hat{X}_{b,t,n} \right|,
\end{equation}
where $X$ is the ground truth, and $\hat{X}$ is the prediction, $B,H,N$ denotes batch size, prediction horizon and traffic stations numbers, respectively. To mitigate expert collapse, we introduce a coefficient-of-variation  based load-balancing loss, $\mathcal{L}_{\text{balance}}$ to encourages global utilization of all heterogeneous experts. 
\begin{equation}
\mathcal{L}_{\text{balance}} = \left( I_{\text{std}} / (\bar{I} + \epsilon \right)^2,
\label{eq:balance_loss}
\end{equation}
where $\mathbf{I} = [I_1,\dots, I_E]$, $I_{\text{std}}$ and $\bar{I}$ denote the standard deviation and mean value,  $\epsilon=10^{-8}$ ensures numerical stability, and $I_i = \sum_{b,t,n} W_{b,t,n,i}$ is the total importance assigned to the $e$-th expert.
The load-balancing loss ensures global diversity, but it does not prevent ambiguous, near-uniform gating. To address this, we incorporate an entropy regularization term, $\mathcal{L}_{\text{entropy}}$, that minimizes the entropy of the gating distribution:
\begin{equation}
    \mathcal{L}_{\text{entropy}} = -\frac{1}{B H N} \sum_{b,t,n} \sum_{e=1}^{E} W_{b,t,n,e} \log(W_{b,t,n,e} + \epsilon).
\label{eq:entropy_loss}
\end{equation}
The entropy loss encourages confident and sharp expert assignment, whereas the load-balancing loss promotes global expert utilization. Their combination balances local routing confidence and global expert diversity.

\section{EXPERIMENTS}
In this section, we conduct extensive experiments to evaluate STHMoE by addressing the following research questions: \textbf{RQ1:} How does STHMoE perform against spatio-temporal graph models and LLM-based baselines across traffic datasets and prediction horizons? \textbf{RQ2:} Are the time, frequency, spatial, and high-order spatial experts, as well as the MoE routing strategy, necessary and effective? \textbf{RQ3:} Do the adaptive structural enhancement module, LLM module, and loss function contribute to performance gains? \textbf{RQ4:} How do key factors, including hypergraph order, loss weights, LLM backbone, and trainable parameters, affect accuracy ? \textbf{RQ5:} What are the computational efficiency and cost of STHMoE?

\subsection{Experimental Setup.}

\paragraph{Datasets.}
We conduct extensive experiments on several real-world traffic datasets, such as traffic flow, index and speed types. The traffic flow datasets~\cite{yu2018spatio} include BIKE-Inflow, PEMS03, PEMS04, PEMS07, and PEMS08. The traffic index and speed datasets~\cite{OpenCity} comprise SZDIDI, CDDIDI,  TrafficHZ, TrafficJN and TrafficNJ. All datasets are partitioned into train/validation/test sets following a split ratio of 7:1:2. Detailed statistics of  datasets are summarized in Table~\ref{datasets1}.

\begin{table}[htbp]
\fontsize{6}{7.2}\selectfont 
\centering
\caption{STATISTICAL DETAILS OF TRAFFIC DATASETS.}
\label{datasets1}
\setlength{\tabcolsep}{1.pt} 

\begin{tabular}{lccccc}
\toprule
Data & Category  & Location  & Regions & Sampling interval (min) & Time span \\ 
\midrule
BIKE & Bike Demand  & New York, USA & 295  & 5    & 2023/01/01—2024/01/01 \\
PEMS03  & Traffic flow & California, USA  & 358  & 5  & 2008/01/01—2008/03/31 \\
PEMS04  & Traffic flow & California, USA  & 307  & 5  & 2018/01/01—2020/02/28 \\
PEMS07  & Traffic speed & California, USA  & 883  & 5  & 2017/05/01—2017/08/31 \\
PEMS08  & Traffic flow & California, USA  & 170  & 5  & 2016/07/01—2020/08/31 \\
SZ DIDI  & Traffic index & Shenzhen, China  & 627  & 10 & 2018/01/01—2018/02/28 \\
CD DIDI  & Traffic index & Chengdu, China  & 524  & 10 & 2018/01/01—2018/02/28 \\
TrafficHZ & Traffic speed & Hangzhou, China  & 672  & 30 & 2022/03/05—2022/04/05 \\
TrafficJN & Traffic speed & Jinan, China   & 576  & 30 & 2022/03/05—2022/04/05 \\
TrafficNJ & Traffic speed & Nanjing, China  & 768  & 30 & 2022/03/05—2022/04/05 \\
\bottomrule
\end{tabular}
\end{table}

\begin{table*}[htbp]
\centering
\fontsize{5.6}{6.72}\selectfont 
\setlength{\tabcolsep}{1.pt}
\renewcommand{\arraystretch}{1.04}
\caption{Performance comparison reformatted from the $(12 \to 12)$, $(24 \to 24)$ and $(48 \to 48)$ settings. IMP.(\%) denotes the relative improvement of STHMoE over the best competing baseline. Negative values indicate that STHMoE is not the best-performing method.}
\label{tab:mainresults}
 
\begin{tabular}{cc|cc|cc|cc|cc|cc|cc|cc|cc|cc|cc|cc|cc|cc|cc|cc}
\hline
\multicolumn{2}{c|}{Model}  & \multicolumn{2}{c|}{IMP.}
& \multicolumn{2}{c|}{\makecell{STHMoE\\(ours)}}
& \multicolumn{2}{c }{\makecell{MSHLLM \\(2026)}}
& \multicolumn{2}{c }{\makecell{SELLM \\(2026)}}
& \multicolumn{2}{c }{\makecell{FactoST \\(2025)}}
& \multicolumn{2}{c }{\makecell{UniST \\ (2025)}}
& \multicolumn{2}{c }{\makecell{STHSepNet\\ (2025)}}
& \multicolumn{2}{c }{\makecell{OpenCity \\ (2025)}}
& \multicolumn{2}{c }{\makecell{STLLM \\ (2025)}}
& \multicolumn{2}{c }{\makecell{FreqMoE \\(2025)}}
& \multicolumn{2}{c}{\makecell{TIMELLM \\ (2024)}}
& \multicolumn{2}{c}{\makecell{LLM4TS \\ (2024)}}
& \multicolumn{2}{c }{\makecell{GPT4TS \\(2023)}}
& \multicolumn{2}{c}{\makecell{TimesNet \\(2023)}}
& \multicolumn{2}{c}{\makecell{STGCN \\(2018)}}  \\
 
\multicolumn{2}{c|}{Dataset} & {MAE} & {RMSE}
& {MAE} & {RMSE}
& MAE & RMSE
& MAE & RMSE
& MAE & RMSE
& MAE & RMSE
& {MAE} & {RMSE}
& MAE & RMSE
& {MAE} & {RMSE}
& MAE & RMSE
& MAE & RMSE
& MAE & RMSE
& MAE & RMSE
& MAE & RMSE
& MAE & RMSE \\
\hline

\multirow{3}*{\rotatebox{90}{BIKE}} & 12 & {+5.64\%} & {+2.25\%} & \textbf{4.52} & \textbf{11.73} & 5.25 & 13.96 & 11.35 & 28.43 & 5.02 & 13.18 & 8.24 & 14.24 & 4.81 & 12.98 & 6.45 & 17.44 & 8.76 & 18.81 & 7.44 & 19.78 & 6.52 & 15.17 & 8.75 & 22.73 & 7.13 & 14.25 & \underline{4.79} & \underline{12.00} & 4.97 & 13.46 \\
 & 24 & {+4.68\%} & {+5.19\%} & \textbf{4.68} & \textbf{12.42} & 6.15 & 16.58 & 8.75 & 22.02 & \underline{4.91} & 13.67 & 8.12 & 15.27 & 5.18 & \underline{13.10} & 8.03 & 18.53 & 7.77 & 17.82 & 5.89 & 13.76 & 6.72 & 16.03 & 5.87 & 15.48 & 7.20 & 16.52 & 5.16 & 14.10 & 5.39 & 14.47 \\

 & 48 & {+0.18\%} & {+3.81\%} & \textbf{5.53} & \textbf{14.40} & 6.45 & 15.50 & 8.88 & 22.32 & 5.74 & 15.02 & 7.90 & 15.96 & 5.69 & \textbf{14.40} & 9.51 & 19.19 & 7.28 & 16.74 & 5.96 & \underline{14.97} & 6.81 & 16.72 & 5.73 & 15.16 & 7.24 & 17.20 & \underline{5.54} & 15.41 & 7.08 & 15.72 \\

\midrule

\multirow{3}*{\rotatebox{90}{PEMS03}} & 12 & {+7.20\%} & {+3.94\%} & \textbf{14.83} & \textbf{25.33} & 35.63 & 49.73 & 30.01 & 47.08 & 17.54 & 28.10 & 40.39 & 53.44 & \underline{15.98} & \underline{26.37} & 17.90 & 28.80 & 18.32 & 29.14 & 21.57 & 34.29 & 21.09 & 36.44 & 22.28 & 35.37 & 22.65 & 39.14 & 23.24 & 37.98 & 16.45 & 27.70 \\
 & 24 & {+3.69\%} & {+3.88\%} & \textbf{16.97} & \textbf{28.50} & 33.56 & 56.40 & 50.04 & 75.56 & 20.32 & 32.58 & 36.30 & 51.57 & \underline{17.62} & \underline{29.65} & 23.39 & 37.92 & 21.39 & 34.89 & 28.83 & 47.01 & 26.23 & 42.59 & 29.17 & 46.65 & 28.77 & 47.31 & 20.07 & 33.38 & 18.02 & 30.59 \\
 & 48 & {+7.70\%} & {+6.15\%} & \textbf{19.41} & \textbf{32.07} & 30.83 & 48.90 & 82.35 & 116.04 & 23.53 & 37.78 & 32.62 & 49.77 & \underline{21.03} & \underline{34.17} & 33.85 & 47.14 & 25.92 & 41.13 & 42.34 & 67.71 & 32.62 & 49.77 & 42.88 & 67.30 & 33.50 & 51.00 & 37.54 & 62.99 & 26.02 & 35.44 \\

\midrule

\multirow{3}*{\rotatebox{90}{PEMS04}} & 12 & {+5.64\%} & {+3.44\%} & \textbf{19.90} & \textbf{32.01} & 42.56 & 64.19 & 31.99 & 47.38 & 23.93 & 37.44 & 42.76 & 59.07 & \underline{21.09} & \underline{33.15} & 24.78 & 40.41 & 22.87 & 34.40 & 27.58 & 42.91 & 27.46 & 43.51 & 32.17 & 49.08 & 29.49 & 46.73 & 29.87 & 46.46 & 22.49 & 34.65 \\
 & 24 & {+1.02\%} & {+0.41\%} & \textbf{21.25} & \textbf{33.88} & 40.90 & 63.79 & 62.80 & 89.65 & 25.45 & 39.40 & 36.59 & 52.68 & \underline{21.47} & \underline{34.02} & 32.97 & 47.49 & 22.29 & 34.66 & 37.63 & 58.76 & 25.88 & 40.18 & 45.16 & 68.47 & 37.12 & 58.29 & 24.45 & 38.92 & 21.58 & 34.26 \\
 & 48 & {+0.83\%} & {+2.62\%} & \textbf{22.76} & \textbf{36.11} & \underline{22.95} & \underline{37.08} & 103.79 & 138.71 & 27.07 & 41.46 & 31.31 & 46.99 & 29.53 & 44.78 & 38.95 & 51.73 & 29.82 & 45.13 & 55.84 & 84.92 & 24.40 & 37.10 & 71.58 & 104.09 & 25.23 & 38.25 & 45.46 & 67.99 & 23.96 & 37.86 \\

\midrule

\multirow{3}*{\rotatebox{90}{PEMS07}} & 12 & {+2.83\%} & {+1.74\%} & \textbf{21.63} & \textbf{34.43} & 39.62 & 50.18 & 45.39 & 66.80 & 26.48 & 41.92 & 40.77 & 54.86 & 22.42 & 35.48 & 44.43 & 65.47 & 24.38 & 37.50 & 31.13 & 47.96 & 29.35 & 46.35 & 33.46 & 49.89 & 31.52 & 49.78 & 33.98 & 52.65 & \underline{22.26} & \underline{35.04} \\
 & 24 & {+1.96\%} & {+0.21\%} & \textbf{23.45} & \textbf{37.78} & 36.73 & 69.68 & 74.92 & 107.28 & 28.60 & 45.40 & 38.16 & 55.29 & \underline{23.92} & \underline{37.86} & 46.72 & 63.15 & 24.41 & 38.88 & 28.83 & 47.01 & 31.61 & 66.97 & 44.91 & 67.64 & 43.10 & 67.58 & 28.65 & 46.72 & 24.70 & 38.81 \\
 & 48 & {+0.39\%} & {+2.53\%} & \textbf{25.42} & \underline{41.45} & 34.05 & 96.75 & 120.99 & 163.21 & 30.89 & 49.16 & 35.72 & 55.72 & \underline{25.52} & \textbf{40.40} & 47.59 & 61.34 & 34.02 & 53.51 & 55.84 & 84.92 & 34.05 & 96.75 & 64.65 & 94.07 & 36.72 & 83.17 & 34.87 & 56.75 & 29.38 & 46.63 \\

\midrule

\multirow{3}*{\rotatebox{90}{PEMS08}} & 12 & {+5.77\%} & {+3.67\%} & \textbf{16.48} & \textbf{25.72} & 35.96 & 44.09 & 31.99 & 47.38 & 18.94 & 29.59 & 35.70 & 46.74 & \underline{17.49} & 26.99 & 32.16 & 48.47 & 22.52 & 31.87 & 31.13 & 47.96 & 22.64 & 34.57 & 24.90 & 37.37 & 24.32 & 37.13 & 24.95 & 38.53 & 17.69 & \underline{26.70} \\
 & 24 & {+0.23\%} & {+0.11\%} & \textbf{17.28} & \underline{27.21} & 37.46 & 48.77 & 53.69 & 76.84 & 21.24 & 32.84 & 33.76 & 46.02 & \underline{17.32} & \textbf{27.18} & 33.83 & 48.67 & 19.59 & 30.24 & 42.89 & 66.97 & 31.92 & 46.29 & 33.40 & 50.19 & 30.17 & 48.20 & 20.10 & 32.58 & 18.59 & 28.71 \\
 & 48 & {+16.08\%} & {+15.85\%} & \textbf{19.05} & \textbf{29.94} & 40.47 & 58.12 & 89.25 & 120.82 & 25.83 & 39.33 & 29.87 & 44.57 & 28.46 & 42.81 & 37.16 & 49.07 & 27.70 & 40.72 & 61.48 & 92.75 & 50.47 & 69.72 & 49.83 & 72.22 & 42.82 & 57.91 & 29.17 & 44.72 & \underline{22.70} & \underline{35.58} \\

\midrule

\multirow{3}*{\rotatebox{90}{SZDIDI}} & 12 & {+0.47\%} & {+0.91\%} & \textbf{2.10} & \textbf{3.27} & 3.05 & 4.71 & 3.34 & 5.08 & 3.76 & 4.17 & 4.15 & 5.94 & 2.67 & 3.74 & 4.30 & 6.16 & 2.20 & 3.40 & 2.92 & 4.54 & 3.01 & 4.85 & 2.97 & 4.58 & 2.84 & 4.39 & 2.33 & 3.73 & \underline{2.11} & \underline{3.30} \\
 & 24 & {+0.93\%} & {+1.20\%} & \textbf{2.12} & \textbf{3.30} & 3.36 & 5.08 & 4.31 & 6.23 & 3.32 & 4.01 & 3.68 & 5.12 & \underline{2.14} & \underline{3.34} & 3.21 & 4.46 & 2.27 & 3.58 & 3.45 & 5.27 & 3.56 & 5.07 & 3.43 & 5.16 & 3.48 & 5.35 & 2.68 & 4.55 & 2.17 & 3.42 \\
 & 48 & {+2.20\%} & {+3.06\%} & \textbf{2.22} & \textbf{3.48} & 2.62 & 4.12 & 4.87 & 6.90 & 2.71 & 3.83 & 3.13 & 4.34 & 2.98 & 4.17 & 3.89 & 4.78 & 2.36 & 3.73 & 3.45 & 5.24 & 3.71 & 5.49 & 3.70 & 5.42 & 3.92 & 5.74 & 2.44 & 3.98 & \underline{2.27} & \underline{3.59} \\

\midrule

\multirow{3}*{\rotatebox{90}{CDDIDI}} & 12 & {+5.26\%} & {+4.66\%} & \textbf{2.34} & \textbf{3.48} & 2.50 & \underline{3.65} & 4.01 & 5.76 & 4.43 & 5.72 & 4.60 & 6.31 & \underline{2.47} & 3.67 & 4.77 & 6.55 & 2.49 & 3.77 & 3.40 & 5.06 & 3.15 & 4.67 & 3.39 & 4.96 & 3.38 & 5.02 & 2.62 & 3.94 & 2.65 & 3.93 \\
 & 24 & {+6.27\%} & {+4.05\%} & \textbf{2.39} & \textbf{3.55} & 2.73 & 4.00 & 5.40 & 7.38 & 3.68 & 4.66 & 3.96 & 5.28 & 2.61 & \underline{3.70} & 4.31 & 5.82 & \underline{2.55} & 3.80 & 4.02 & 5.89 & 3.87 & 5.70 & 3.94 & 5.62 & 4.20 & 6.19 & 2.60 & 3.92 & 2.67 & 3.89 \\
 & 48 & {+6.82\%} & {+6.32\%} & \textbf{2.46} & \underline{3.64} & 2.96 & 4.20 & 6.89 & 8.93 & \underline{2.64} & \textbf{3.41} & 3.05 & 3.87 & 2.90 & 3.71 & 3.79 & 4.26 & 2.67 & 3.98 & 4.02 & 5.75 & 4.75 & 6.75 & 4.31 & 5.95 & 4.94 & 6.97 & 2.82 & 4.16 & 2.97 & 3.76 \\

\midrule

\multirow{3}*{\rotatebox{90}{TrafficJN}} & 12 & {+7.55\%} & {+4.35\%} & \underline{0.53} & 1.06 & 0.60 & 1.12 & 0.85 & 1.67 & 0.67 & \textbf{1.18} & 1.21 & 1.39 & \textbf{0.49} & \underline{0.92} & 0.70 & 0.95 & 0.56 & 1.03 & 0.80 & 1.57 & 0.76 & 1.46 & 0.83 & 1.63 & 0.81 & 1.57 & 0.62 & 1.22 & 0.60 & 1.11 \\
 & 24 & {+3.92\%} & {+3.03\%} & \textbf{0.49} & 1.00 & 0.56 & 1.05 & 0.83 & 1.62 & 0.58 & \textbf{1.09} & 0.84 & 1.27 & \underline{0.51} & 1.01 & 0.68 & 1.09 & 0.53 & \underline{0.99} & 0.73 & 1.48 & 0.67 & 1.42 & 0.84 & 1.65 & 0.79 & 1.52 & 0.58 & 1.13 & \underline{0.51} & 1.01 \\
 & 48 & {+6.52\%} & {+1.09\%} & \textbf{0.43} & \underline{0.92} & 0.47 & 0.95 & 0.71 & 1.38 & 0.47 & 1.05 & 0.54 & 1.18 & 0.48 & 0.96 & 0.67 & 1.30 & \underline{0.46} & \textbf{0.91} & 0.49 & 0.94 & 0.73 & 1.33 & 0.68 & 1.32 & 0.75 & 1.35 & 0.55 & 1.12 & 0.50 & 1.00 \\

\midrule

\multirow{3}*{\rotatebox{90}{TrafficNJ}} & 12 & {+4.08\%} & {+5.94\%} & \textbf{0.47} & \textbf{0.95} & 0.68 & 1.27 & 1.07 & 2.08 & 0.65 & 1.13 & 1.00 & 1.17 & \underline{0.49} & \underline{1.01} & 1.09 & 1.39 & 1.02 & 2.52 & 0.94 & 1.87 & 0.89 & 1.69 & 0.87 & 1.70 & 0.96 & 1.81 & 0.54 & 1.13 & 0.67 & 1.25 \\
 & 24 & {-2.17\%} & {-1.05\%} & \underline{0.47} & \underline{0.96} & 0.85 & 1.69 & 1.02 & 1.99 & 0.64 & 1.07 & 0.82 & 1.16 & 0.48 & 0.99 & 0.97 & 1.31 & 1.18 & 1.68 & 0.79 & 1.64 & 0.74 & 1.41 & 0.79 & 1.58 & 0.80 & 1.50 & 0.51 & 1.06 & \textbf{0.46} & \textbf{0.95} \\
 & 48 & {+2.27\%} & {0.00\%} & \textbf{0.43} & \textbf{0.91} & \underline{0.44} & \underline{0.92} & 0.88 & 1.65 & 0.62 & 1.01 & 0.71 & 1.14 & 0.45 & 0.93 & 0.89 & 1.26 & 0.68 & 1.10 & 0.45 & 0.96 & 0.56 & 1.04 & 0.51 & 1.03 & 0.58 & 1.06 & 0.50 & 1.07 & 0.45 & \textbf{0.91} \\

\midrule

\multirow{3}*{\rotatebox{90}{TrafficHZ}} & 12 & {+12.90\%} & {-15.25\%} & \textbf{0.27} & \underline{0.68} & 0.32 & 0.73 & 0.45 & 1.01 & 0.43 & 0.69 & 0.75 & 1.10 & 0.32 & 0.78 & 0.33 & \textbf{0.59} & 0.55 & 1.14 & 0.42 & 0.95 & 0.41 & 0.97 & 0.42 & 0.95 & 0.44 & 1.04 & 0.32 & 0.77 & \underline{0.31} & 0.71 \\
 & 24 & {+6.90\%} & {+1.45\%} & \textbf{0.27} & \textbf{0.68} & 0.43 & 0.97 & 0.46 & 0.99 & 0.45 & 0.71 & 0.74 & 1.05 & 0.32 & 0.76 & 0.62 & 0.84 & 0.59 & 0.97 & 0.39 & 0.89 & 0.42 & 0.95 & 0.41 & 0.92 & 0.37 & 0.87 & 0.30 & 0.75 & \underline{0.29} & \underline{0.69} \\
 & 48 & {+7.41\%} & {+2.99\%} & \textbf{0.25} & \textbf{0.65} & 0.28 & \underline{0.67} & 0.38 & 0.86 & 0.62 & 0.87 & 0.72 & 0.99 & 0.31 & 0.76 & 0.90 & 1.09 & 0.69 & 0.95 & 0.28 & 0.69 & 0.44 & 0.89 & 0.31 & 0.73 & 0.45 & 0.91 & 0.31 & 0.79 & \underline{0.27} & 0.69 \\
\midrule
\end{tabular} 
\end{table*}

\paragraph{Baseline Models.} 
To verify the effectiveness of STHMoE, we benchmark it against state-of-the-art methods across three distinct categories as follows:
 \begin{itemize}
\item \textbf{Time Series Forecasting Models:} 
TimesNet~\cite{wu2023timesnet} transforms 1D time series into 2D tensors to capture multi-periodic intra- and inter-period variations. 
FreqMoE~\cite{liu2025freqmoe} employs a frequency-domain Mixture-of-Experts based on distinct frequency characteristics.

\item \textbf{Spatio-Temporal Prediction Models:} 
STGCN~\cite{yu2018spatio} integrates graph convolutions and gated 1D temporal convolutions to capture spatial and temporal dependencies. 
FactoST~\cite{zhong2025learning} factorizes pre-training into two stages, featuring general temporal pre-training followed by spatial-specific adaptation to decouple spatio-temporal representations. 
OpenCity~\cite{OpenCity} combines transformer and graph neural network architectures, leveraging pre-training on large-scale heterogeneous traffic data.

\item \textbf{Large Language Prediction Models:} UniST~\cite{yuan2024unist} utilizes prompt-empowered mechanisms to perform universal urban spatio-temporal forecasting. STHSepNet~\cite{chen2025decoupling} combines a lightweight LLM with an adaptive hypergraph to capture higher-order spatial interactions. STLLM~\cite{liu2025st} introduces a spatio-temporal embedding layer and a partially frozen LLM mechanism. MSHLLM~\cite{shang2026multiscale} aligns multi-scale hypergraph features with LLMs via hybrid prompt guidance. GPT4TS~\cite{zhou2023one} adapts pre-trained language models to time-series forecasting through representation alignment and fine-tuning. LLM4TS~\cite{chang2023llm4ts} aligns time series representations with language model embeddings. TIMELLM~\cite{jin2024time} and SELLM~\cite{li2024gpt23} reprogram numerical time series into textual prototypes to leverage frozen LLMs for prediction. Here, STLLM, MSHLLM, STHSepNet, UniST, and OpenCity are configured with GPT-2 as the pre-trained backbone.

 \end{itemize}

\paragraph{Implementation Details} All experiments are conducted using PyTorch on 2$\times$NVIDIA RTX A6000 GPUs (48GB). We adopt a pre-trained BERT (110M parameters) as the LLM backbone (LoRA rank $r=16$) for fine-tuning. Key structural hyperparameters are set as follows: historical window L={12,24,48}, forecasting horizon H={12,24,48}, MixProp layers $K=2$, and hypergraph order $k=3$. STHMoE is optimized by Adam optimizer (learning rate $0.005$, weight decay $0.0001$, batch size 16). Training proceeds for 20 epochs with a patience of 5 epochs.  

\subsection{Experimental Results \textbf{(RQ1)}}
Table~\ref{tab:mainresults} reports the performance comparison between STHMoE and 13 state-of-the-art baselines on 10 real-world traffic datasets under three forecasting horizons, i.e., 12, 24, and 48 steps. STHMoE achieves the best results in most evaluated settings and remains competitive in the remaining cases, indicating the effectiveness of the proposed framework. 

Across the evaluated datasets, STHMoE generally achieves lower MAE and RMSE than temporal forecasting models, spatio-temporal graph models, and LLM-based baselines. The ``IMP.'' column reports the relative improvement of STHMoE over the best competing baseline; negative values indicate cases where STHMoE is not the best-performing method. Notably, on PEMS08 under the 48-step setting, STHMoE reduces MAE and RMSE by 16.08\% and 15.85\%, respectively. On PEMS03, STHMoE obtains MAE improvements of 7.20\%, 3.69\%, and 7.70\% for the 12-, 24-, and 48-step settings, respectively. These results indicate that the proposed framework can effectively integrate complementary temporal, frequency, and spatial information for traffic forecasting.

For most traffic-flow datasets, prediction errors increase as the forecasting horizon extends, reflecting the growing uncertainty of long-term forecasting. For several traffic speed and traffic index datasets, the trend is less monotonic, suggesting that forecasting difficulty is also affected by data scale, sampling interval, and temporal regularity. Under the 48-step setting, STHMoE achieves notable MAE improvements on PEMS03, PEMS08, CDDIDI, and TrafficJN, with relative gains of 7.70\%, 16.08\%, 6.82\%, and 6.52\%, respectively. These results suggest that expert decomposition and adaptive routing are helpful for long-horizon prediction, although the magnitude of improvement varies across datasets. The evaluated datasets cover diverse traffic scenarios, including highway traffic flow, urban traffic speed, ride-hailing indices, and bike-sharing demand. STHMoE shows favorable performance across these scenarios compared with temporal models, spatio-temporal models, and LLM-based forecasting methods. The gains observed on datasets such as PEMS08, CDDIDI, and TrafficHZ suggest that adaptive higher-order structural modeling and heterogeneous expert fusion can provide useful complementary information when traffic data exhibit complex temporal fluctuations or spatial dependencies.

\begin{table*}[htbp]
\fontsize{5.8}{6.96}\selectfont 
\centering
\caption{Ablation study of domain-specific experts. Performance comparison reformatted from the $(12 \to 12)$ setting.}
\label{table02}
\setlength{\tabcolsep}{1.2pt} 
\begin{tabular}{lcccc|*{21}{c}}
\toprule 
\multicolumn{5}{c|}{Model}   
& \multicolumn{2}{c}{BIKE} 
& \multicolumn{2}{c}{PEMS03} 
& \multicolumn{2}{c}{PEMS04}  
& \multicolumn{2}{c}{PEMS07}
& \multicolumn{2}{c}{PEMS08} 
& \multicolumn{2}{c}{SZDIDI} 
& \multicolumn{2}{c}{CDDIDI} 
& \multicolumn{2}{c}{TrafficJN} 
& \multicolumn{2}{c}{TrafficNJ} 
& \multicolumn{2}{c}{TrafficHZ}
\\
& Frequency & Time & Spatio & Hypergraph    
& MAE & RMSE 
& MAE & RMSE 
& MAE & RMSE  
& MAE & RMSE 
& MAE & RMSE 
& MAE & RMSE 
& MAE & RMSE 
& MAE & RMSE
& MAE & RMSE 
& MAE & RMSE \\
\midrule

Frequency Expert  
& \cmark & \textcolor{red}{\xmark} & \textcolor{red}{\xmark} & \textcolor{red}{\xmark} 
& 4.73 & 12.07 
& 15.61 & 26.26 
& 23.12 & 35.23 
& 21.75 & 34.39 
& 16.91 & 26.08 
& 2.11 & \underline{3.25} 
& 2.49 & 3.66 
& 0.57 & 0.78 
& 0.51 & 1.02 
& \underline{0.28} & 0.71 \\

Time Expert  
& \textcolor{red}{\xmark} & \cmark & \textcolor{red}{\xmark} & \textcolor{red}{\xmark} 
& \underline{4.55} & \underline{11.94} 
& 15.47 & 26.11 
& 20.24 & 32.04 
& 21.90 & \textbf{34.27} 
& \underline{16.86} & 26.06 
& 2.30 & 3.56 
& 2.50 & 3.67 
& 0.57 & \underline{0.76} 
& 0.53 & \textbf{0.73} 
& 0.29 & 0.72 \\

Spatio Expert  
& \textcolor{red}{\xmark} & \textcolor{red}{\xmark} & \cmark & \textcolor{red}{\xmark} 
& 4.70 & 12.25 
& 15.55 & 26.61 
& 20.55 & 32.22 
& 21.69 & \underline{34.32} 
& 16.96 & 25.99 
& 2.38 & 3.51 
& 2.47 & 3.64 
& 0.56 & \textbf{0.75} 
& 0.54 & 0.81 
& 0.29 & 0.72 \\

Higher-order Spatio Expert  
& \textcolor{red}{\xmark} & \textcolor{red}{\xmark} & \textcolor{red}{\xmark} & \cmark
& 4.93 & 12.61 
& 16.15 & \underline{25.86} 
& 20.16 & 31.92 
& \underline{21.67} & 34.75 
& 16.88 & \underline{25.78} 
& 2.11 & \underline{3.25} 
& 2.59 & 3.79 
& 0.61 & \underline{0.76} 
& 0.58 & \underline{0.76} 
& 0.30 & 0.72 \\

\midrule 

w/o-Frequency
& \textcolor{red}{\xmark} & \cmark & \cmark & \cmark  
& 4.64 & 12.19 
& \underline{15.18} & 27.89 
& 20.23 & 32.26 
& 22.40 & 34.79 
& 17.16 & 26.16 
& 2.23 & 3.42 
& 2.44 & 3.61 
& 0.59 & 1.13 
& 0.49 & 0.99 
& 0.29 & 0.70 \\

w/o-Time  
& \cmark & \textcolor{red}{\xmark} & \cmark & \cmark  
& 4.63 & 12.01 
& 15.49 & 27.15 
& 20.06 & \underline{31.78} 
& 23.58 & 36.81 
& 17.03 & 26.71 
& 2.13 & 3.27 
& 2.36 & \underline{3.49} 
& 0.58 & 1.11 
& 0.50 & 1.01 
& 0.30 & 0.71 \\

w/o-Spatial  
& \cmark & \cmark & \textcolor{red}{\xmark} & \cmark 
& 4.60 & 12.00 
& 15.97 & 27.84 
& 20.47 & 32.60 
& 22.66 & 35.25 
& 18.20 & 27.93 
& \textbf{2.09} & \textbf{3.24} 
& 2.41 & 3.54 
& 0.59 & 1.15 
& 0.51 & 1.03 
& \underline{0.28} & \underline{0.69} \\

w/o-Higher-order Spatial  
& \cmark & \cmark & \cmark & \textcolor{red}{\xmark}
& 4.67 & 12.03 
& 15.59 & 26.31 
& \underline{19.92} & \textbf{31.64} 
& 22.07 & 34.68 
& 17.13 & 26.46 
& \underline{2.10} & 3.32 
& \underline{2.35} & \textbf{3.48} 
& \textbf{0.48} & 0.97 
& \underline{0.48} & 0.98 
& \underline{0.28} & \underline{0.69} \\

\midrule 

STHMoE  
& \cmark & \cmark & \cmark & \cmark
& \textbf{4.52} & \textbf{11.73} 
& \textbf{14.83} & \textbf{25.33} 
& \textbf{19.90} & 32.01 
& \textbf{21.63} & 34.43 
& \textbf{16.48} & \textbf{25.72} 
& \underline{2.10} & 3.27 
& \textbf{2.34} & \textbf{3.48} 
& \underline{0.53} & 1.06 
& \textbf{0.47} & 0.95 
& \textbf{0.27} & \textbf{0.68} 
\\

\midrule
\multicolumn{5}{c|}{IMP. (\%)} 
& 0.66 & 1.76 
& 2.31 & 2.05 
& 0.10 & -1.17 
& 0.18 & -0.47 
& 2.25 & 0.23 
& -0.48 & -0.93 
& 0.43 & 0.00 
& -10.42 & -41.33 
& 2.08 & -30.14 
& 3.57 & 1.45 \\   
\bottomrule
\end{tabular}
\end{table*}

\subsection{Ablation Studies}

\subsubsection{Ablation of Sub-Experts \textbf{(RQ2)}}
To examine the role of each domain expert, we evaluate two groups of variants. The single-expert variants retain only one of the Frequency, Time, Spatio, and Higher-order spatial experts, whereas the leave-one-out variants remove one expert from the full STHMoE framework. Table~\ref{table02} reports the full model achieves the best or tied-best results on 13 out of 20 metrics, showing more stable performance across datasets than the reduced variants. On PEMS03, STHMoE obtains an MAE of 14.83 and an RMSE of 25.33, improving over the strongest variant by 2.31\% and 2.05\%, respectively. It also performs favorably on BIKE and PEMS08, indicating that heterogeneous expert fusion is generally more effective than relying on a single representation.

The single-expert variants show that each individual expert captures only part of the forecasting information. For example, the Frequency Expert reports an MAE of 23.12 on PEMS04, which is 16.18\% higher than that of STHMoE. The spatio expert yields a 5.05\% higher RMSE on PEMS03, and the Higher-order spatio expert increases the MAE on BIKE from 4.52 to 4.93. These results indicate that frequency-domain, temporal, and spatial cues are complementary rather than individually sufficient. The leave-one-out results provide a similar observation. Removing the Time Expert increases the RMSE on PEMS07 by 6.91\%, while removing the Frequency Expert raises the PEMS07 MAE from 21.63 to 22.40. The w/o-Spatial variant leads to a 10.44\% MAE increase on PEMS08, suggesting the contribution of pairwise spatial modeling. Removing the Hypergraph Expert increases the TrafficNJ RMSE from 0.95 to 0.98, indicating that higher-order spatial interactions also provide useful information. These results support the benefit of integrating temporal, frequency, spatial, and higher-order spatial experts in STHMoE.

\subsubsection{Ablation of MOE~\textbf{(RQ2)}}
\begin{figure}[htbp]
    \centering
    \includegraphics[width=0.95\linewidth]{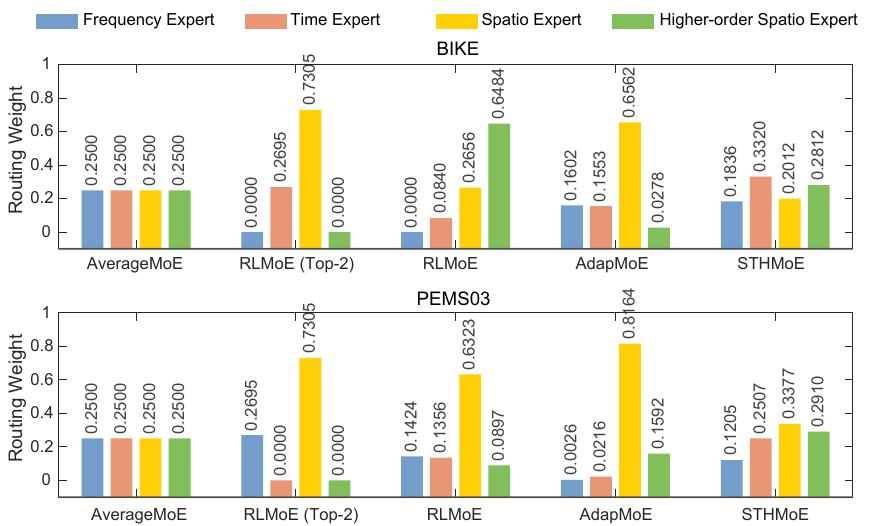}
    \caption{Comparison of routing weight distributions (expert utilization distributions) under different MoE fusion mechanisms}
    \label{fig:moe}
\end{figure}

To evaluate the effect of the expert fusion strategy, we compare STHMoE with several MoE variants, including AverageMoE, RLMoE (Top-2), RLMoE, and AdapMoE. All variants use the same expert set, namely the Frequency Expert, Time Expert, Spatio Expert, and Higher-order Spatio Expert, so that the comparison mainly reflects the effect of the routing mechanism. The results are reported in Table~\ref{tab:moe_fusion}, and the averaged routing weight distributions on BIKE-Inflow and PEMS03 are shown in Fig.~\ref{fig:moe}.

\begin{table}[htbp]
\scriptsize
\centering
\fontsize{6}{7.2}\selectfont 
\setlength{\tabcolsep}{1.2pt} 
\caption{Ablation study on Mixture-of-Experts fusion mechanisms.}
\label{tab:moe_fusion}
\begin{tabular}{l*{12}{c}}
\toprule
Fusion Variant 
& \multicolumn{2}{c}{BIKE} 
& \multicolumn{2}{c}{PEMS03} 
& \multicolumn{2}{c}{PEMS04}  
& \multicolumn{2}{c}{PEMS08} 
& \multicolumn{2}{c}{PEMS-07}
& \multicolumn{2}{c}{SZDIDI}  \\
& MAE & RMSE 
& MAE & RMSE 
& MAE & RMSE 
& MAE & RMSE
& MAE & RMSE 
& MAE & RMSE  \\
\midrule

AverageMoE  
& \underline{4.69} & \underline{11.96}
& \underline{15.97} & \underline{26.22}
& \underline{20.19} & \underline{32.37}
& \textbf{21.61} & 34.72
& 19.13 & \textbf{25.18}
& \underline{2.36} & \underline{3.49} \\

RLMoE (Top) 
& 4.95 & 13.27 
& 20.35 & 31.35 
& 27.25 & 36.91 
& 25.73 & \textbf{33.98} 
& 23.35 & 32.40 
& 2.48 & 3.85 \\

RLMoE 
& 4.88 & 11.99 
& 18.37 & 29.85 
& 23.40 & 36.91 
& 25.05 & 39.58 
& 18.70 & 28.70 
& 2.43 & 3.72 \\

AdapMoE   
& 5.71 & 13.22 
& 17.36 & 29.80 
& 23.41 & 36.80 
& 24.07 & 35.27 
& \underline{18.63} & 28.20 
& 2.43 & 3.74 \\

\midrule

STHMoE (Ours)  
& \textbf{4.52} & \textbf{11.73} 
& \textbf{14.83} & \textbf{25.33} 
& \textbf{19.90} & \textbf{32.01}  
& \underline{21.63} & \underline{34.43} 
& \textbf{16.48} & \underline{25.72} 
& \textbf{2.10} & \textbf{3.27} \\

\midrule
IMP. (\%) 
& 3.62 & 1.92
& 7.14 & 3.39
& 1.44 & 1.11
& -0.09 & -1.32
& 11.54 & -2.14
& 11.02 & 6.30 \\
\bottomrule
\end{tabular}
\end{table}

Table~\ref{tab:moe_fusion} shows that STHMoE achieves the best results on 9 out of 12 metrics and remains close to the best fusion variant in the remaining cases. Compared with AverageMoE, which is the strongest overall competing fusion variant in this comparison, STHMoE reduces MAE/RMSE by 3.62\%/1.92\% on BIKE and by 7.14\%/3.39\% on PEMS03. These results show that coordinate-wise routing provides a more flexible fusion strategy than fixed averaging when combining heterogeneous expert outputs. Fig.~\ref{fig:moe} further illustrates the routing behavior of different MoE variants. RLMoE and AdapMoE assign relatively large weights to a limited number of experts, while other experts receive smaller contributions. In contrast, STHMoE produces a less concentrated routing distribution and shows dataset-dependent expert preferences. For example, it assigns relatively higher weights to temporal information on BIKE-Inflow, whereas spatial and spatio-temporal experts receive larger weights on PEMS03. This observation  suggests that the proposed routing strategy can better adapt expert contributions to different traffic patterns.

\begin{table}[htbp]
\fontsize{6}{7.2}\selectfont 
\centering
\caption{Ablation Study of Key Components in STHMoE. DEGR.(\%) denotes the relative performance degradation of each ablated variant compared with the STHMoE full model.}
\label{ablationconponents}
\setlength{\tabcolsep}{1pt}
\begin{tabular}{lcccccccccccc}
\toprule
\multirow{2}{*}{} 
& \multicolumn{2}{c}{BIKE} 
& \multicolumn{2}{c}{PEMS03} 
& \multicolumn{2}{c}{PEMS04}   
& \multicolumn{2}{c}{PEMS08} 
& \multicolumn{2}{c}{SZDIDI} 
& \multicolumn{2}{c}{CDDIDI} \\
& MAE & RMSE 
& MAE & RMSE 
& MAE & RMSE 
& MAE & RMSE 
& MAE & RMSE 
& MAE & RMSE \\
\midrule

STHMoE 
& 4.52 & 11.73 
& 14.83 & 25.33 
& 19.90 & 32.01 
& 16.48 & 25.72 
& 2.10 & 3.27 
& 2.34 & 3.48 \\

\midrule 

w/o-structure  
& 4.79 & 12.47 
& 15.36 & 25.44 
& 20.57 & 32.39 
& 16.67 & 26.02 
& 2.11 & 3.29 
& 2.43 & 3.57 \\
DEGR.(\%)  
& 5.97 & 6.31 
& 3.57 & 0.43 
& 3.37 & 1.19 
& 1.15 & 1.17 
& 0.48 & 0.61 
& 3.85 & 2.59 \\

\midrule 

w/o-llm 
& 4.65 & 12.08
& 15.31 & 25.76 
& 20.09 & 31.76 
& 17.94 & 27.55 
& 2.38 & 3.52
& 2.39 & 3.61
\\

DEGR.(\%)  
& 2.88 & 2.98
& 3.24 & 1.70
& 0.95 & -0.78
& 8.86 & 7.12
& 13.33 & 7.65
& 2.14 & 3.74 \\

\midrule 

w/o-balance loss  
& 4.71 & 12.02 
& 15.29 & 25.96 
& 19.72 & 31.92 
& 16.71 & 25.78 
& 2.18 & 3.34 
& 2.35 & 3.49 \\
DEGR.(\%)  
& 4.20 & 2.47 
& 3.10 & 2.49 
& -0.90 & -0.28 
& 1.40 & 0.23 
& 3.81 & 2.14 
& 0.43 & 0.29 \\

\midrule 

w/o-entropy loss 
& 4.74 & 12.25 
& 15.59 & 26.47 
& 20.40 & 32.22 
& 16.63 & 25.66 
& 2.17 & 3.32 
& 2.34 & 3.47 \\
DEGR.(\%)  
& 4.87 & 4.43 
& 5.12 & 4.50 
& 2.51 & 0.66 
& 0.91 & -0.23 
& 3.33 & 1.53 
& 0.00 & -0.29 \\

\bottomrule 
\end{tabular}
\end{table}

\subsubsection{Ablation Study of Core Components~\textbf{(RQ3)}}
To evaluate the contribution of the key components, we ablate the adaptive structure-enhanced module, the LLM backbone, and the two routing regularization terms. Table~\ref{ablationconponents} shows that removing these components weakens forecasting performance, although the degree of degradation varies across datasets and metrics.

\paragraph{Structure-Enhanced Module (w/o-structure)} The removal of the adaptive structure modeling module degrades predictive performance across all datasets. The largest drop is observed on BIKE, where MAE and RMSE increase by 5.97\% and 6.31\%, respectively. This result indicates that the learned first-order adjacency matrix and higher-order hypergraph provide useful structural information for modeling dynamic spatial dependencies and group-wise interactions.

\paragraph{LLM Backbone (w/o-llm)} The absence of the large language model backbone severely deteriorates overall forecasting accuracy. Specifically, the model's MAE on the SZDIDI and PEMS08 datasets increases by 13.33\% and 8.86\%, respectively. This suggests that the pre-trained LLM, together with lightweight adaptation, provides useful transferable representations for modeling temporal patterns in traffic sequences.

\paragraph{Load-Balancing Loss (w/o-balance loss)} Without the auxiliary load-balancing loss, prediction errors increase on most evaluated datasets. For example, the MAE on BIKE and PEMS03 rises by 4.20\% and 3.10\%, respectively. This result suggests that load balancing helps prevent over-concentration on a small number of experts and promotes more stable utilization of heterogeneous experts during training.

\paragraph{Entropy Regularization Loss (w/o-entropy loss)}
The exclusion of the entropy regularization loss compromises model accuracy, increasing MAE and RMSE on the PEMS03 dataset by 5.12\% and 4.50\%, respectively. This result indicates that entropy regularization helps the router produce more confident local expert assignments, reducing the ambiguity caused by overly uniform expert weights. Together with load balancing, it contributes to a better trade-off between global expert utilization and local routing confidence.

\begin{table}[htbp]
\centering
\scriptsize
\caption{Comparison among LLM-based baselines and our
framework across different LLM backbones.  
}

\label{tab:llmbackbone}
\fontsize{5.4}{6.48}\selectfont 
\setlength{\tabcolsep}{1pt}
\begin{tabular}{clcccccccccccccc}
\toprule
\multicolumn{2}{c|}{Backbone}  & \multicolumn{2}{c|}{IMP.(\%)} & \multicolumn{2}{c}{\makecell{STHMoE\\ (ours)}} & \multicolumn{2}{c}{\makecell{STLLM\\ (2025)}} & \multicolumn{2}{c}{\makecell{STHSepNet\\ (2025)}} 
& \multicolumn{2}{c}{\makecell{TIMELLM \\ (2024)}}   
& \multicolumn{2}{c}{\makecell{GPT4TS \\(2023)}}  
& \multicolumn{2}{c}{\makecell{MSHLLM \\(2026)}}     \\

\multicolumn{2}{c|}{Dataset} &   MAE   & RMSE    &  MAE   & RMSE   & MAE   & RMSE  & MAE     & RMSE    & MAE    & RMSE   & MAE    & RMSE   & MAE    & RMSE     \\

\midrule

\multirow{4}{*}{\rotatebox{90}{BiKE}} 

& BERT   & +5.44 & +9.63
&  \textbf{4.52}     &  \textbf{11.73}       
&     8.76  &  18.81           
&   4.81      &      \underline{12.98}       
&   6.52  &  15.17      
&  \underline{4.78}   &    13.32
&  5.25  &   13.96
\\

& GPT2   & +8.13 & +7.46
&   \textbf{4.63}    &    \textbf{12.04}           
&  8.47   &  18.03     
&     \underline{5.04} &  \underline{13.01}   
&  8.55  & 21.03   
& 6.63 & 17.53 
& 11.38 & 28.44
\\

& LLaMA-1B     & +11.50 & +8.79
&  \textbf{4.85}   &  \textbf{13.60}   
&  8.36 &   18.44  
&  \underline{5.48}   &   \underline{14.91}  
&  8.41   &   20.58  
&   6.96  &   18.37 
& 11.37 & 28.45
\\

& DeepSeek-1.5B & -21.91 & -30.85
&  \underline{5.62}    
& \underline{15.61}   
&  8.79   &   18.98
& \textbf{4.61}  &  \textbf{11.93}  
&  8.38 & 20.52    
& 6.65 & 17.36   
& 11.28 & 27.69
\\

\midrule

\multirow{4}{*}{\rotatebox{90}{PEMS03}} 

& BERT   & +7.20 & +3.94 
&   \textbf{14.83}     &   \textbf{25.33}     
&   18.32 & 29.14  
& \underline{15.98}    &     \underline{26.37}    
&  21.09      &  36.44     
&   22.65  & 39.14   
&  35.63  &  49.73      
\\

& GPT2   & +1.10 & -2.22   
&   \textbf{15.34}  &   \underline{26.74}  
&   20.53  &   32.55
&   \underline{15.51}  & \textbf{26.16}      
& 31.84   &  48.70 
& 21.14 &  33.89  
& 30.15 & 46.89  
\\

& LLaMA-1B     & +5.62 & +0.52   
&  \textbf{15.79} &  \textbf{26.95}   
&  29.96 &  47.02 
&  \underline{16.73}   & \underline{27.09}    
&  46.50  &  68.92   
&  21.38  & 34.09 
& 29.94 &  47.01  
\\

& DeepSeek-1.5B & +2.90 & +0.04
&  \textbf{15.73}  &  \textbf{26.83} 
& 29.99  & 47.01   
& \underline{16.20} &  \underline{26.84}    
&     43.50  &  61.87   
&     21.32  &      33.89  
&  30.07 &  46.90
\\

\midrule

\multirow{4}{*}{\rotatebox{90}{PEMS04}} 

& BERT   & +5.64 & +3.44 
&  \textbf{19.90}     &  \textbf{32.01}      
&       22.87  & 34.40    
& \underline{21.09}   &   \underline{33.15}    
&     27.46  &     43.51 
&       29.49  &  46.73  
& 42.56  &  64.19   
\\

& GPT2   & +0.25 & -1.24   
&  \textbf{20.29}    &  \underline{32.71}    
&  22.35  &    34.96    
&     \underline{20.34}  &  \textbf{32.31}  
&  50.09  &  71.57  
&  27.48 &  43.19 
&  38.48 & 56.60   
\\

& LLaMA-1B     & +6.52 & +9.69
&  \textbf{22.07}     & \textbf{34.12}   
& 38.35  & 56.67 
&  \underline{23.61}   & \underline{37.78}     
&  57.03  & 75.39   
&  27.99  &    43.23
& 38.47 &   56.63 
\\

& DeepSeek-1.5B & +3.55 & +1.01
& \textbf{21.20}  &  \textbf{33.41}  
& 38.35  &  56.67   
&   \underline{21.98} &      \underline{33.75}
&  43.50  &   61.87
&    27.48  &   43.19 
& 38.46  & 56.61
\\

\midrule

\multirow{4}{*}{\rotatebox{90}{PEMS07}} 

& BERT   & +3.52 & +2.96   
&  \textbf{21.63}  & \textbf{34.43}     
&  24.38  &  37.50   
&   \underline{22.42} &   \underline{35.48}      
&   29.35 &  46.35  
&  31.52 &  49.78 
&  39.62 &  50.18   
\\

& GPT2     & +17.00 & +17.07     
&  \textbf{22.07} &  \textbf{34.68}  
& 30.54  & 47.36  
& \underline{26.59}    & \underline{41.82}    
& 66.05   & 90.01  
& 30.82  & 48.02   
& 45.56 & 66.61 
\\

& LLaMA-1B     & +20.24 & +19.81   
&  \textbf{22.66} & \textbf{35.25}   
&  45.32 & 66.81  
& \underline{28.41}  & \underline{43.96}   
& 70.78   &   95.97 
& 32.48 & 49.03  
& 47.52    &  69.06  
\\

& DeepSeek-1.5B & +13.71 & +15.46  
& \textbf{23.72}  &  \textbf{36.10}  
&   45.38 & 66.80   
& \underline{27.49}    &  \underline{42.70}   
& 68.71   & 92.86   
& 30.82 &   48.03 
& 45.58  &  66.59 
\\

\midrule

\multirow{4}{*}{\rotatebox{90}{PEMS08}} 

& BERT   & +5.77 & +4.71   
&   \textbf{16.48}  &  \textbf{25.72}    
&      22.52     & 31.87      
& \underline{17.49}  &  \underline{26.99}   
&   22.64      &   34.57   
&  24.32  &   37.13  
&    35.96  &  44.09  
\\

& GPT2   & +9.31 & +5.61     
&    \textbf{16.17}    & \textbf{25.41}   
&  25.37  & 36.92  
&     \underline{17.83}    &  \underline{26.92}
&       48.08 &  64.86    
&  22.11  &    34.92 
& 32.20 & 47.32  
\\

& LLaMA-1B     & +8.53 & +6.20     
&  \textbf{16.83} &  \textbf{25.87}  
&  31.99   & 47.38  
& \underline{18.40}  &    \underline{27.58} 
&    53.69   &  71.70 
&22.53  & 35.04 
&  32.19 &  47.32  
\\

& DeepSeek-1.5B & -2.31 & -4.69     
& \underline{17.28}  &  \underline{27.21}  
& 31.97  & 47.39  
&  \textbf{16.89}   &  \textbf{25.99}   
&  49.35  &67.07   
&   22.10  &  34.92  
&  32.15 &  47.30  
\\

\bottomrule

\end{tabular}

\end{table}

\begin{figure}[htbp]
    \centering
    \includegraphics[width=1\linewidth]{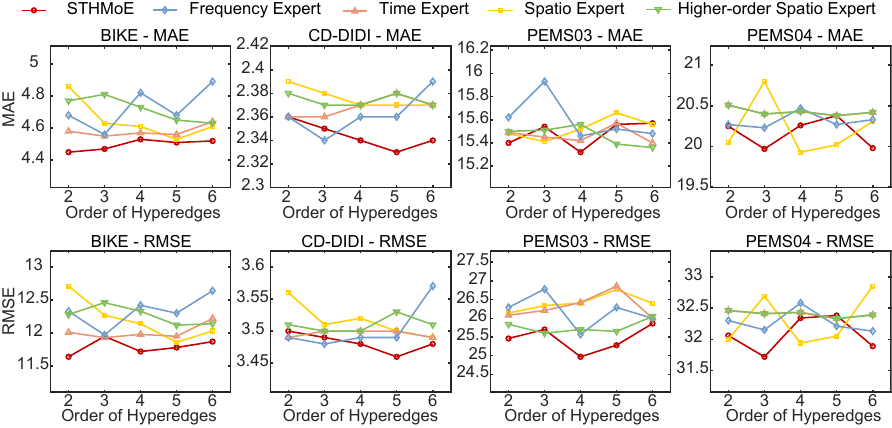}
\caption{Effect of hyperedge order ($k$ in Eq.~\eqref{eq:higherorder}) on the MAE and RMSE of STHMoE and its experts, with $k \in \{2,3,4,5,6\}$.}
    \label{fig:higherorder}
\end{figure}

\subsubsection{Comparison across Different LLM Backbones. \textbf{(RQ4)}}
To further investigate whether the performance gains of STHMoE stem from specific large language models or from the proposed spatio-temporal decoupling and adaptive spatial modeling, we conduct a comprehensive
comparison across multiple LLM backbones.
Specifically, we evaluate TIMELLM, STLLM, STHMoE, GPT4TS, STHSepNet and MSHLLM under 
identical experimental settings using different pre-trained models, including BERT, GPT2, LLaMA-1B, and DeepSeek-1.5B.

Table~\ref{tab:llmbackbone} shows STHMoE achieves favorable results on most dataset-backbone combinations. With BERT as the backbone, STHMoE obtains the best performance on the evaluated BIKE and PEMS datasets, suggesting that the proposed framework can work well with a moderate-size pre-trained model. Similar trends are observed with GPT-2 and LLaMA-1B in many cases. For example, on PEMS07, STHMoE achieves positive MAE improvements across all four backbones, with gains ranging from 3.52\% to 20.24\%. On PEMS08, STHMoE also performs competitively under BERT, GPT-2, and LLaMA-1B. Meanwhile, the results indicate that adopting a larger or more recent LLM backbone does not necessarily guarantee better forecasting accuracy. Under DeepSeek-1.5B, STHMoE is slightly inferior to STHSepNet on BIKE and PEMS08, resulting in negative IMP values.

\subsection{Parameter Analysis}

\paragraph{Impact of hyperedge order \textbf{(RQ4)}}
Fig.~\ref{fig:higherorder} shows the sensitivity of STHMoE and its expert variants to the hyperedge order $k$. As $k$ varies from 2 to 6, both MAE and RMSE change non-monotonically across datasets, indicating that larger hyperedges do not necessarily lead to better forecasting accuracy. A moderate order can capture informative group-wise spatio-temporal dependencies, whereas an excessively large order may introduce redundant or noisy relations. Compared with individual expert variants, STHMoE shows more stable performance in most settings, suggesting that MoE-based fusion can better coordinate frequency-domain, temporal, spatial, and higher-order spatial information. These results highlight the importance of selecting an appropriate hyperedge order for effective higher-order spatial modeling.

\begin{figure}[htbp]
\centering
\includegraphics[width=1\linewidth]{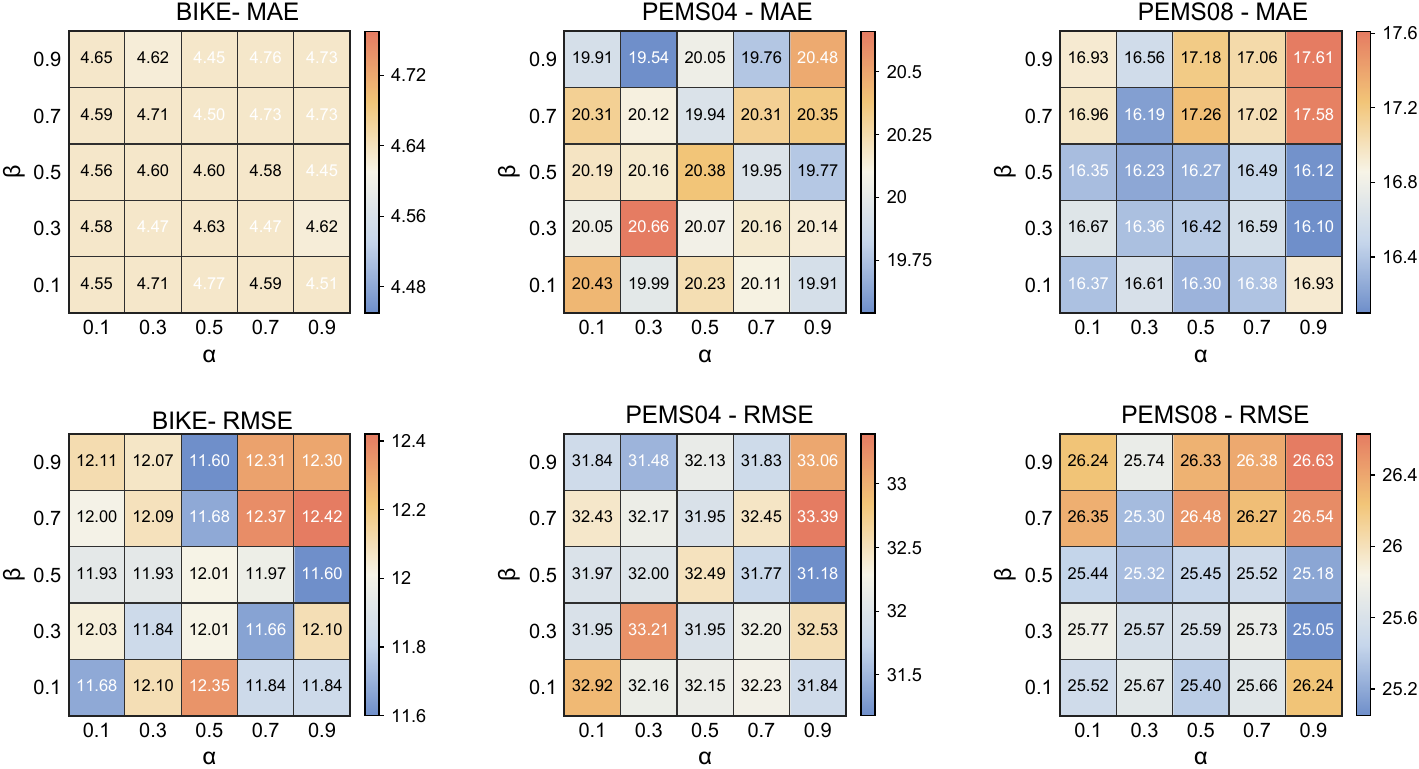}
\caption{Effect of balance loss and entropy loss weights ($\alpha$ and $\beta$ in Eq.~\eqref{eq:total_loss}) on model performance. MAE and RMSE are reported on BIKE-Inflow, PEMS04, and PEMS08.}
\label{fig:lossfunction}
\end{figure}

\paragraph{Loss function \textbf{(RQ4)}}
Fig.~\ref{fig:lossfunction} reports the sensitivity of STHMoE to the balance loss weight $\alpha$ and entropy loss weight $\beta$ in Eq.~\eqref{eq:total_loss}. Both weights affect MAE and RMSE, indicating the importance of routing regularization. In general, a relatively large $\alpha$ yields more stable performance, suggesting that load balancing helps prevent expert collapse and improves global expert utilization. Meanwhile, the entropy term is also sensitive to its weight: a moderate $\beta$ encourages more confident expert assignment, whereas an overly large value may over-sharpen the routing distribution and interfere with the forecasting objective. For example, $\alpha=0.9$ with a moderate $\beta$ provides robust performance across multiple datasets, and $\alpha=0.9,\beta=0.3$ achieves the best results on PEMS08. These results suggest that load-balancing and entropy regularization play complementary roles, balancing global expert utilization with local routing confidence.

\subsection{Complexity Analysis \textbf{(RQ5)}}

\begin{figure}[htbp]
\centering
\includegraphics[width=1\linewidth]{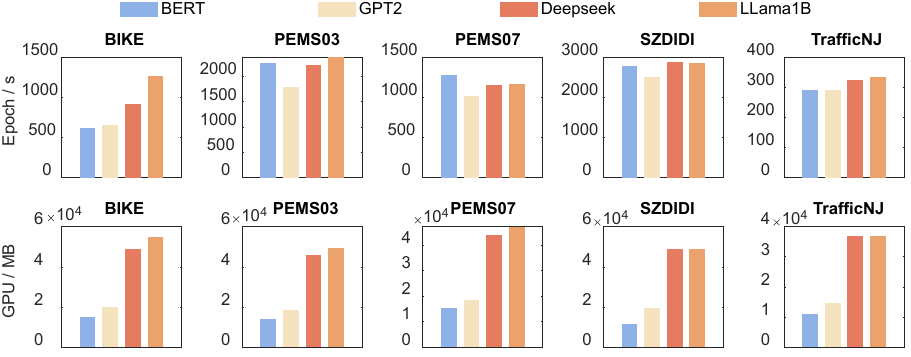}
\caption{Training time (s/epoch) and peak GPU memory (MB) of STHMoE with different LLM backbones. The average epoch time and peak GPU memory usage are reported across traffic datasets.}
\label{fig:timegpu}
\end{figure}

Fig.~\ref{fig:timegpu} reports the average epoch time and GPU memory consumption of STHMoE with different LLM backbones. On  training time, GPT-2 achieves the shortest average epoch time across the five datasets, reducing the epoch time by 13.63\%, 16.84\%, and 22.10\% compared with BERT, DeepSeek-1.5B, and LLaMA-1B, respectively. More specifically, GPT-2 decreases the training time on PEMS03, PEMS07, SZDIDI by 20.93\%, 20.75\%, and 10.18\% relative to BERT, while slightly increasing the runtime on BIKE by 6.33\%. For memory consumption, BERT exhibits the best GPU memory efficiency. On average, BERT requires 26.46\% less memory than GPT-2, 70.09\% less than DeepSeek, and 71.61\% less than LLaMA-1B. These results suggest that BERT is more suitable for memory-constrained deployment scenarios, whereas GPT-2 offers faster training in most evaluated cases.

\section{Conclusion}
 
In this paper, we proposed STHMoE, a spatio-temporal hypergraph-enhanced Mixture-of-Experts framework for traffic forecasting. STHMoE addresses heterogeneous dependency coordination in urban traffic big data by organizing temporal, spectral, pairwise spatial, and higher-order structural cues into specialized LLM-based experts and adaptively fusing them at fine-grained spatio-temporal coordinates. To model evolving spatial dependencies, an adaptive structure-enhanced module is introduced to learn both first-order graph relations and higher-order hypergraph interactions. An entropy-aware MoE router further aggregates expert outputs at fine-grained spatio-temporal coordinates while balancing expert utilization and routing confidence. Experiments on ten real-world traffic datasets show that STHMoE achieves competitive performance against representative time-series, spatio-temporal graph, and LLM-based baselines. Ablation and sensitivity studies further support the contribution of expert decomposition, adaptive structural modeling, the LLM backbone, and routing regularization. Future work will explore more efficient routing strategies, the incorporation of multi-source external factors, and more interpretable expert specialization for large-scale spatio-temporal data mining.

\bibliographystyle{IEEEtran} 
\bibliography{New_IEEEtran}

\end{document}